\documentclass[10pt, journal, twocolumn]{IEEEtran}

\usepackage{color,graphicx,amsmath,amssymb,amsthm,epsfig,mathrsfs,cite,bm,graphics}
\usepackage{caption}\usepackage{algpseudocode}
\usepackage{algorithmicx}
\usepackage{algorithm} 
\usepackage[utf8]{inputenc}
\usepackage{amsmath}
\usepackage{amsfonts}
\usepackage{amssymb}
\usepackage{placeins}
\usepackage{graphicx}
\usepackage{bbm}
\usepackage{graphicx,graphics,subcaption}
\usepackage{url}

\usepackage{tikz}
\usetikzlibrary{automata, positioning, arrows,shapes.multipart}

\usepackage{amsmath,amsthm,bm}
\usepackage{amssymb}
\newcommand{\norm}[1]{\left\lVert#1\right\rVert}
\usepackage{xcolor}
\usepackage{graphics,graphicx,color,epsfig}
\usepackage{tikz}
\usetikzlibrary{automata, positioning, arrows,shapes.multipart,fit}
\usepackage{pgfplots}
\usepackage{lipsum}
\usepackage{mathtools}
\usepackage{cuted}
\usepackage{hyperref}

\newcommand{\imp}[1]{#1}
\newcommand{\added}[1]{#1}
\newcommand{\revise}[1]{#1}
\newcommand{\cutt}[1]{}
\newcommand{\cut}[1]{}
\usepackage{multirow}

\newtheorem{theorem}{Theorem}[]
\newtheorem{lemma}[]{Lemma}

\newtheorem{definition}[]{Definition}

\title{Tune smarter not harder: A principled approach to tuning learning rates for shallow nets}

\author{ Thulasi Tholeti* \thanks{*The authors are with the Department of Electrical Engineering, Indian Institute of Technology Madras, Chennai, India 600 036. Email: \texttt{\{ee15d410,skalyani\}@ee.iitm.ac.in}} \hspace{16pt} Sheetal Kalyani*
}

\begin{document}
    \newcommand\copyrighttext{
	\Huge {IEEE Copyright Notice} \\ \\
	\large {Copyright © 2020 IEEE \\
		Personal use of this material is permitted. Permission from IEEE must be obtained for all other uses, in any current or future media, including reprinting/republishing this material for advertising or promotional purposes, creating new collective works, for resale or redistribution to servers or lists, or reuse of any copyrighted component of this work in other works.} 
		
		\vspace{2cm}
	
	{\Large Published in IEEE Transactions on Signal Processing (Volume: 68) on  26 August 2020.  DOI: 10.1109/TSP.2020.3019655. Link found \href{https://ieeexplore.ieee.org/document/9178494}{here}. } \\ \\ 
	
	\vspace{2cm}
	
	Cite as:\\
	
    \fbox{\begin{minipage}{25em}
    T. Tholeti and S. Kalyani, "Tune Smarter Not Harder: A Principled Approach to Tuning Learning Rates for Shallow Nets," in IEEE Transactions on Signal Processing, vol. 68, pp. 5063-5078, 2020, doi: 10.1109/TSP.2020.3019655.
    \end{minipage}}

}

\twocolumn[
\begin{@twocolumnfalse}
	\copyrighttext
\end{@twocolumnfalse}
]

	\maketitle
    \begin{abstract}
    Effective hyper-parameter tuning is essential to guarantee the performance that neural networks have come to be known for. In this work, a principled approach to choosing the learning rate is proposed for shallow feedforward neural networks. We associate the learning rate with the gradient Lipschitz constant of the objective to be minimized while training. An upper bound on the mentioned constant is derived and a search algorithm, which always results in non-divergent traces, is proposed to exploit the derived bound. It is shown through simulations that the proposed search method significantly outperforms the existing tuning methods such as Tree Parzen Estimators (TPE). \revise{The proposed method is applied to three different existing applications: a) channel estimation in OFDM systems, b) prediction of the exchange currency rates and c) offset estimation in OFDM receivers}, and it is shown to pick better learning rates than the existing methods using the same or lesser compute power.
\end{abstract}

\section{Introduction}
Deep neural networks have made significant improvements to fields like speech and image processing \cite{litjens2017survey}, communications \cite{mao2018deep,raj2018backpropagating,raj2020design}, computer vision, etc. \cite{liu2017survey}. These networks are typically trained using an iterative optimization algorithm such as Gradient Descent (GD) or its multiple variants \cite{yu1995adaptive,kingma2014adam}. To successfully deploy these networks for various applications, the hyper-parameters of the network, namely the width and the depth of the network and the learning rate used for training should be carefully tuned \cite{feurer2019hyperparameter}.

Initially, manual search and grid search were the most popular approaches \cite{montgomery2001design}. The authors of \cite{bergstra2012random} then showed that randomly chosen trials were more efficient in terms of search time for hyper-parameter optimization than a grid-based search. However, in both the methods, the observations from the previous samples are not utilized to choose values for the subsequent trials. To remedy this, Sequential Model-Based Optimization (SMBO) was introduced to perform hyper-parameter tuning where the next set of hyper-parameters to be evaluated are chosen based on the previous trials \cite{snoek2012practical}. 
Some of the well-known models for Bayesian optimization are Gaussian Processes \cite{rasmussen2003gaussian}, random forests \cite{hutter2011sequential} and TPE \cite{bergstra2013making}. 

In the methods listed here so far, the tuning of hyper-parameters is typically performed as a black-box module, i.e., without utilizing any information about the objective function to be minimized. There exist many applications in which the architecture of the network is fixed, for which the number of layers and the width of the network are already specified and are not treated as hyper-parameters. Given such an architecture, the learning rate is an important hyper-parameter as it determines the speed of convergence of the optimization algorithm \cite{senior2013empirical}. In such cases, it would be beneficial if the learning rate is derived as a function of the objective as it can be simply recomputed for a new set of inputs instead of tuning the learning rate from scratch. \\
\revise{\indent The idea of tuning-free algorithms has recently attracted attention, not only in neural networks but in the context of other algorithms as well. For example, \cite{kallummil2018signal} proposed a tuning-free Orthogonal Matching Pursuit (OMP) algorithm, \cite{chaudhuri2009parameter} proposed a tuning-free hedge algorithm and \cite{menon2019structured} proposed a parameter-free robust Principal Component Analysis (PCA) method.} To propose such a tuning-free equivalent for the GD algorithm while training neural networks, it would require a theoretical analysis of the objective function. Although neural networks are applied to varied applications, little is known about its theoretical properties when the network consists of multiple hidden layers. Most theoretical works such as \cite{zhang2019learning,li2017convergence} are available for networks with one or two hidden layers, which we call shallow networks.\\
\indent Although deep neural networks are popular in computer vision and image processing where the objective function is complex, applications in areas like wireless communication and finance predictions still employ shallow feedforward neural networks as evidenced by works in \cite{jiang2019deep,tacspinar2010back,galeshchuk2016neural,huang2018deep,ye2017neural}. In \cite{tacspinar2010back}, channel estimation for Orthogonal Frequency Division Multiplexing (OFDM) systems was done using a single hidden layer network. Shallow networks were also used in applications like user equipment localization \cite{ye2017neural}, symbol detection in high-speed OFDM underwater acoustic communication \cite{chen2018neural} and Direction of Arrival (DoA) estimation \cite{huang2018deep}. In all the above, the architecture for a given application was fixed and the learning rate was chosen by manual tuning or grid search.\\
\indent For such applications which employ a fixed shallow architecture, a theory-based approach for choosing the learning rate will save the computation which would otherwise be spent on tuning hyper-parameters. The learning rate of the optimization algorithm has been associated with the Lipschitz properties of the objective function, namely the Lipschitz constant of the gradient of the objective function in \cite{nesterov1998introductory}. Although, there has been significant interest in analyzing the Lipschitz properties of neural networks in recent literature \cite{balan2017lipschitz,virmaux2018lipschitz}, these works focus on the Lipschitz constant of the output which plays an important role in analysing the stability of the network, and not on the gradient Lipschitz constant of the objective which is required for quantifying the learning rate. 
\subsection{Motivation}
In the existing works on hyper-parameter optimization, the choice of learning rate is often treated as a separate module that is to be performed before the training; they do not employ any information about the function that should be optimized. As an alternative, we wish to associate the learning rate with the parameters of the problem, thereby providing a theoretical justification to the choice of learning rate and also use this to tune in a smarter fashion. 
In typical tuning methods, there is a clear trade-off between the number of trials of the search algorithm that is allowed and the performance of the chosen learning rate. If one decides to adopt a higher number of trials then, one is more likely to achieve a better learning rate. However, there is no guarantee that the chosen learning rate will lead to convergent behaviour of GD given any fixed number of trials. In the proposed method, we wish to provide the user with the same trade-off between the number of trials and the performance, whilst ensuring that chosen learning rate always results in convergence irrespective of the number of trials allowed. \cut{This is achieved as the search algorithm uses as its starting point the derived learning rate, which is guaranteed to converge, as it is an upper bound on the gradient Lipschitz constant.}
 
\vspace{-0.3cm}
\subsection{Contributions}
A theory-based approach to determine the learning rate for shallow networks is proposed. The contributions of this work are four-fold. Firstly, using classic literature \cite{nesterov1998introductory}, the learning rate is associated with the gradient Lipschitz constant of the objective function. \cut{We propose to use this property to set the learning rate of the training algorithm instead of a black-box search.} Secondly, the upper bound on gradient Lipschitz constants for feedforward neural networks consisting of one and two layers are derived for popular activation functions, namely, ReLU and sigmoid. The bounds, initially in terms of eigenvalues of large Hessian matrices, are simplified to yield easy-to-implement expressions that can be adapted to a given architecture. Thirdly, the derived bound on the gradient Lipschitz constant is utilized for determining the learning rate; an algorithm, 'BinarySearch', is introduced for this search. The proposed algorithm is shown to outperform the popular hyper-parameter tuning estimator, TPE, in terms of the loss achieved, \added{while ensuring convergence.} \revise{Finally, the utility of the proposed method is also demonstrated using three applications: channel estimation in the case of OFDM systems, Carrier Frequency Offset estimation in OFDM receivers and the prediction of exchange rates for currencies.} 



\vspace{-0.4cm}
\subsection{Notation}
We use bold upper-case letters, say \(\bm{A}\) to denote matrices and \(A_{ij},\bm{A}^i\) to denote their \((i,j)_{th}\) element and the \(i_{th}\) column respectively. The maximum eigenvalue of \(\bm{A}\) is denoted as \(\lambda_{max}(\bm{A})\); the maximum diagonal entry is denoted as \(\mathcal{D}_{max}(\bm{A})\). The bold lower-case letters \(\bm{x},\bm{y}\) denote vectors.  All vectors are column vectors unless stated otherwise. The \(\ell_2 \) norm of a vector is denoted as \(\norm{.}\). The \(\ell_1 \) and \(\ell_{\infty}\) norms of a vector \(\bm{x}\) are denoted as \(|\bm{x}|_1= \sum_{i} x_i\) and \(|\bm{x}|_{\infty} = \max_{i} x_i\) respectively. The indicator function denoted as \(\mathbb{I}_{\mathcal{E}}\) takes the value 1 when \(\mathcal{E}\) is true and value \(0\) otherwise. The symbols \(\nabla\) and \(\nabla^2\) denote the first and second derivatives respectively.


    \vspace{0.1cm}
\section{Definitions and Background} \label{sec:motiv}
\noindent
\begin{definition}
A differentiable function \(f: \mathbb{R}^d \rightarrow \mathbb{R}\) is said to be \(\alpha\)- gradient Lipschitz if for any \(\bm{x_1},\bm{x_2}\) in the domain of \(f\), and for \(\alpha > 0\),
\begin{equation} \label{eqn:lipschitz}
    \norm{ \nabla f(\bm{x_1}) - \nabla f(\bm{x_2}) } \leq \alpha \norm{\bm{x_1} - \bm{x_2} },
\end{equation}
where \(\alpha\) is known as the gradient Lipschitz constant. The smallest such constant is known as the optimal constant, denoted by \(\alpha^*\).
\end{definition} 

 Nesterov's seminal work \cite{nesterov1998introductory} discusses the following theorem which guarantees the convergence of the GD algorithm. 
\begin{lemma}\cite{nesterov1998introductory} \label{lemma:descent}
    For an \(\alpha\)-gradient Lipschitz function \(f: \mathbb{R}^d \rightarrow \mathbb{R}\), gradient descent with a step size \(\eta \leq 1/\alpha\) produces a decreasing sequence of objective values and the optimal step size is given by \(\eta^* = 1/\alpha\).
\end{lemma}
For a doubly differentiable function \(f\) with gradient Lipschitz constant as \(\alpha\), we have \cite{nesterov1998introductory} 
\begin{equation} \label{eqn:condition}
    \nabla^2 f(\bm{x}) \preceq \alpha \bm{I} \quad \forall x.
\end{equation}
This implies that all eigenvalues of the matrix \(\nabla^2 f(\bm{x})-\alpha \bm{I}\) should be less than or equal to zero for all values of \(\bm{x}\). This is achieved when the maximum eigenvalue satisfies this condition. Therefore, the gradient Lipschitz constant of a double differentiable function is given by 
\begin{equation} \label{eqn:maxEig}
     \alpha^* = \max_{\bm{x}} \lambda_{max}(\nabla^2 f(\bm{x})).
\end{equation}
We use (\ref{eqn:maxEig}) in the following sections to derive the required constant.
Note that any \(\alpha > \alpha^*\) also satisfies (\ref{eqn:condition}). Therefore, if the exact value for \(\alpha^*\) cannot be determined, an upper bound on \(\alpha^*\) can be derived. The learning rate derived from the upper bound also results in a decreasing sequence of iterates according to Lemma \ref{lemma:descent}. This signifies that the learning rate derived as the inverse of the gradient Lipschitz constant or any upper bound will always result in convergence of the gradient descent algorithm. This implication is used by us to guarantee the convergence of GD while training neural networks.

	\section{Deriving the gradient Lipschitz constant for a single hidden layer neural network} \label{sec:1layer}
In this section, a neural network with a single hidden layer consisting of \(k\) neurons with activation function \(act(.)\) is considered, as given in Fig. \ref{fig:1Larch}. We derive the gradient Lipschitz constant for two different popular activation functions: sigmoid and ReLU. 
The weight vector from the input to the \(j\)th hidden layer neuron is denoted as \(\bm{w}^j\) where \(\bm{w}^j \in \mathbb{R}^d\) for \(j = 1,...k\). The column vector \(\bm{w}\) refers to the stack of vectors \(\bm{w}^1,...\bm{w}^k\); \(\bm{w} \in \mathbb{R}^{kd}\). The output of the network is taken as the sum of outputs from each of the hidden layer neurons and is given by \(f(\bm{x},\bm{w}) = \sum_{j=1}^k act(\bm{x}^T \bm{w}^j)\) for input \(\bm{x}\). 
The training data is denoted as a set of points \((\bm{x}(i),y(i))\)
for \(i = 1,...N\). The aim of the network is to learn the function \(f\) given the training data. Throughout, we consider the quadratic loss function namely,
    \begin{equation} \label{eqn:loss1L}
              l(\bm{w}) = \dfrac{1}{2N} \sum_{i = 1}^N \left(  \left(\sum_{j=1}^k act(\bm{x}(i)^T \bm{w}^j) \right) - y(i) \right)^2.
     \end{equation}

\tikzset{%
  every neuron/.style={
    circle,
    draw,
    minimum size=1cm
  },
  neuron missing/.style={
    draw=none, 
    scale=4,
    text height=0.333cm,
    execute at begin node=\color{black}$\vdots$
  },
}
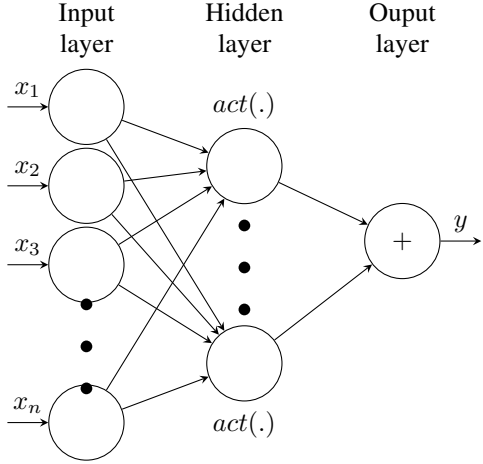
\begin{figure}
    \centering
\begin{tikzpicture}[scale = 0.7,x=1.5cm, y=1.5cm, >=stealth]

\foreach \m/\l [count=\y] in {1,2,3,missing,4}
  \node [every neuron/.try, neuron \m/.try] (input-\m) at (0,2.5-\y) {};

\foreach \m [count=\y] in {1,missing,2}
  \node [every neuron/.try, neuron \m/.try ] (hidden-\m) at (2,2-\y*1.25) {};

\foreach \m [count=\y] in {1}
  \node [every neuron/.try, neuron \m/.try ] (output-\m) at (4,-0.2) {$+$};

\foreach \l [count=\i] in {1,2,3,n}
  \draw [<-] (input-\i) -- ++(-1,0)
    node [above, midway] {$x_\l$};

\foreach \l [count=\i] in {1}
  \node [above] at (hidden-\i.north) {$act(.)$};
  
\foreach \l [count=\i] in {2}
  \node [below] at (hidden-\l.south) {$act(.)$};

\foreach \l [count=\i] in {1}
  \draw [->] (output-\i) -- ++(1,0)
    node [above, midway] {$y$};

\foreach \i in {1,...,4}
  \foreach \j in {1,...,2}
    \draw [->] (input-\i) -- (hidden-\j);

\foreach \i in {1,...,2}
  \foreach \j in {1}
    \draw [->] (hidden-\i) -- (output-\j);

\foreach \l [count=\x from 0] in {Input, Hidden, Ouput}
  \node [align=center, above] at (\x*2,2) {\l \\ layer};

\end{tikzpicture}
    \caption{The architecture of a single hidden layer network}
    \label{fig:1Larch}
\end{figure}

\vspace{-0.3cm}
\subsection{Sigmoid activation}
The sigmoid activation is defined as \(\sigma(x) = \frac{1}{1 + \exp(-x)}. \) The gradient Lipschitz constant for a single hidden layer network with sigmoid activation function is derived in this section.
    Initially, we consider a single data point, \((\bm{x},y)\), and then extend it to a database. 
     
     \begin{theorem}\label{theorem:optConvSig}
     The gradient Lipschitz constant for a single-hidden layer feedforward network with sigmoid activation when considering quadratic loss function in (\ref{eqn:loss1L}) with \(act(.)=\sigma(.)\) and \(N=1\) is given by,
          \begin{equation}
          \alpha^* \leq \min \bigg(\frac{|k-y|}{10} + \frac{k}{16}, 0.1176 (k-1)  + \frac{|y|}{10} +0.077\bigg) \norm{\bm{x}}^2 .
      \end{equation}
     \end{theorem}
     \begin{proof}
      As the loss function is doubly differentiable, the required constant is \(\alpha^* = \max_{\bm{w}} \lambda_{max}(\nabla^2 l(\bm{w}) )\). Note,
      \begin{equation}
         \nabla l(\bm{w}) = \bigg( \sum_{j=1}^k \sigma(\bm{x}^T \bm{w}^j) - y \bigg) 
         \begin{bmatrix} 
         \sigma(\bm{x}^T \bm{w^1}) (1 - \sigma(\bm{x}^T \bm{w^1})) \bm{x}\\ 
         \vdots \\ 
         \sigma(\bm{x}^T \bm{w^k}) (1 - \sigma(\bm{x}^T \bm{w^k})) \bm{x} \end{bmatrix}.
    \end{equation}
    \begin{equation}
        \bm{\hat{b}(x,w)} \triangleq \begin{bmatrix} 
         \sigma(\bm{x}^T \bm{w^1}) (1 - \sigma(\bm{x}^T \bm{w^1}))\\ 
         \vdots \\ 
         \sigma(\bm{x}^T \bm{w^k}) (1 - \sigma(\bm{x}^T \bm{w^k})) \end{bmatrix},
    \end{equation}
is defined where
    \(\bm{\hat{b}(x,w)} \in \mathbb{R}^{k}\). Let \(Diag_m(k_m)\) denote a diagonal matrix whose non-zero entry in the \(m_{th}\) row is \(k_m\). The Hessian matrix computed using the product rule of differentiation is given by, 
     \begin{align}
         \nabla^2 l(\bm{w}) &= \Bigg(Diag_m \Bigg(
         \big( \sum_{j=1}^k \sigma(\bm{x}^T \bm{w}^j) - y \big) \sigma(\bm{x}^T \bm{w}^m)  \nonumber  \\ 
         &\hspace{1cm} (1 - \sigma(\bm{x}^T \bm{w}^m)) (1 - 2 \sigma(\bm{x}^T \bm{w}^m)) \Bigg)  \nonumber \\
         &\hspace{1cm} + \bm{\hat{b}(x,w)}  \bm{\hat{b}(x,w)}^T \Bigg) \otimes \bm{x x^T}.
     \end{align}
     The gradient Lipschitz constant is given by
     \begin{align}
         \alpha^* &= \max_{\bm{w}} \lambda_{max}\Bigg[ 
         \Bigg(Diag_m \Bigg(
         \big( \sum_{j=1}^k \sigma(\bm{x}^T \bm{w}^j) - y \big) \sigma(\bm{x}^T \bm{w}^m)  \nonumber \\ 
         &\hspace{1cm} (1 - \sigma(\bm{x}^T \bm{w}^m)) (1 - 2 \sigma(\bm{x}^T \bm{w}^m)) \Bigg)  \nonumber  \\
         &\hspace{1cm} + \bm{\hat{b}(x,w)}  \bm{\hat{b}(x,w)}^T \Bigg) \otimes \bm{x x^T} 
         \Bigg]. \label{eqn:eqn27}
     \end{align}
     Note (\ref{eqn:eqn27}) involves a maximization over all possible values of \(\bm{w}\) and an eigenvalue computation for every value. We use the structure of the matrix to provide a simplified solution. We use the following property of Kronecker products \cite{laub2005matrix}.
     \begin{lemma} \label{lemma:kron}
        Let \(\bm{A} \in \mathbb{R}^{n \times n}\) have eigenvalues \(\lambda_i, i \in n\), and let \(\bm{B} \in \mathbb{R}^{m \times m}\) have eigenvalues \(\mu_j, j \in m\), then the \(mn\) eigenvalues of \(\bm{A} \otimes \bm{B}\) are given by 
        \begin{equation*}
            \lambda_1\mu_1, \cdots, \lambda_1\mu_m, \lambda_2\mu_1, \cdots, \lambda_2\mu_m, \cdots, \lambda_n\mu_m.
        \end{equation*}
     \end{lemma}
      Therefore, the maximum eigenvalue of the Kronecker product will be the product of the maximum eigenvalues, if the maximum eigenvalue of the diagonal matrix is positive; else, it will be zero. As we are maximizing over all possible values of \(\bm{w}\), we can always ensure that the maximum eigenvalue is positive. Since \(\bm{xx^T}\) is rank one with a single non-zero eigenvalue, \(\bm{x^Tx}\), using Lemma \ref{lemma:kron}, we have,
      \begin{align} \label{eqn:eqn32}
          \alpha^* = \max_{\bm{w}}  \lambda_{max}(\bm{P}) \bm{x^Tx} ,
      \end{align}
      where \(\bm{P}\) is defined as
     \begin{align}
         \bm{P} &\triangleq Diag_m \Bigg(
         \big( \sum_{j=1}^k \sigma(\bm{x}^T \bm{w}^j) - y \big) \sigma(\bm{x}^T \bm{w}^m)  \nonumber \\ 
         &\hspace{1cm} (1 - \sigma(\bm{x}^T \bm{w}^m)) (1 - 2 \sigma(\bm{x}^T \bm{w}^m)) \Bigg)  \nonumber  \\
         &\hspace{1cm} + \bm{\hat{b}(x,w)}  \bm{\hat{b}(x,w)}^T.
     \end{align}
     A bound can be obtained to find the maximum eigenvalue of \(\bm{P}\) using the Weyl's inequality which states that for Hermitian matrices \(\bm{A}\) and \(\bm{B}\),
     \begin{equation} \label{eqn:weyl}
        \lambda_{\max} (\bm{A}+\bm{B}) \leq \lambda_{\max}(\bm{A}) + \lambda_{\max}(\bm{B}).
    \end{equation} 
    Using the above inequality, the observation that \(\bm{\hat{b}(x,w)}  \bm{\hat{b}(x,w)}^T\) is rank-1 and that the eigenvalues of a diagonal matrix are the diagonal entries, one obtains,  
    \begin{align}
         \lambda_{max}(\bm{P}) &\leq \max_m \Bigg(
         \big( \sum_{j=1}^k \sigma(\bm{x}^T \bm{w}^j) - y \big) \sigma(\bm{x}^T \bm{w}^m)  \nonumber \\ 
         &\hspace{1cm} (1 - \sigma(\bm{x}^T \bm{w}^m)) (1 - 2 \sigma(\bm{x}^T \bm{w}^m)) \Bigg)  \nonumber  \\
         &\hspace{1cm} + \bm{\hat{b}(x,w)}^T  \bm{\hat{b}(x,w)}. \label{eqn:eqn33}
     \end{align}
    Combining (\ref{eqn:eqn32}) and (\ref{eqn:eqn33}),
    \begin{align}
         \alpha^* 
          & \leq \max_{\bm{w}} \Bigg(\max_m \bigg(
         \big( \sum_{j=1}^k \sigma(\bm{x}^T \bm{w}^j) - y \big)\nonumber \\ 
         &\hspace{1cm} \sigma(\bm{x}^T \bm{w}^m) 
         (1 - \sigma(\bm{x}^T \bm{w}^m)) (1 - 2 \sigma(\bm{x}^T \bm{w}^m)) \bigg) \nonumber \\
         &\hspace{1cm} +  \norm{\bm{\hat{b}(x,w)}}^2 \Bigg)  \norm{\bm{x}}^2.
         \label{eqn:29}
     \end{align}
     The expression in (\ref{eqn:29}) can be written in terms of the derivatives of sigmoid function as given below:
     \begin{align}
         \alpha^* 
          & \leq \max_{\bm{w}} \Bigg(\max_m \bigg(
         \big( \sum_{j=1}^k \sigma(\bm{x}^T \bm{w}^j) - y \big)\nonumber \\ 
         &\hspace{0.5cm} \nabla^2 \sigma(\bm{x}^T \bm{w}^m) 
          \bigg) +  \sum_{j=1}^k \big(\nabla \sigma(\bm{x}^T \bm{w}^j) \big)^2 \Bigg)  \norm{\bm{x}}^2. \label{eqn:derivative}
     \end{align}
     We now use the following bounds on the sigmoid derivatives  \cite{schlessman2002approximation} to bound (\ref{eqn:derivative}):
     \begin{equation} \label{eqn:func}
             0 \leq \sigma(x) \leq 1 \quad \forall x
         \end{equation}
        \begin{equation}\label{eqn:func1stder}
             \nabla_x \sigma(x) = \sigma(x)(1 - \sigma(x)) \leq \frac{1}{4} \quad \forall x
         \end{equation}
        \begin{equation}\label{eqn:func2ndder}
             \nabla_x^2 \sigma(x) = \sigma(x)(1 - \sigma(x))(1 - 2\sigma(x)) \leq \frac{1}{10} \quad \forall x.
         \end{equation}
     Using the above conditions to individually maximize each of the terms in (\ref{eqn:derivative}),
     \begin{equation} \label{eqn:bound1sig}
         \alpha^* \leq  \left[\frac{|k-y|}{10} + \frac{k}{16} \right]  \norm{\bm{x}}^2.
     \end{equation}
     We note that tighter bounds may be achieved by maximizing (\ref{eqn:derivative}) as a whole instead of each individual term. As the maximization in (\ref{eqn:derivative}) is over the weights \(\bm{w}\), considering the terms consisting of \(\bm{w}\), 
     \begin{align}\label{eqn:termswithw}
         & \max_m \bigg(
         \big( \sum_{j=1}^k \sigma(\bm{x}^T \bm{w}^j) - y \big) \nabla^2 \sigma(\bm{x}^T \bm{w}^m) 
          \bigg) +  \nonumber \\ 
         &\hspace{3cm} \sum_{j=1}^k \big(\nabla \sigma(\bm{x}^T \bm{w}^j) \big)^2
     \end{align}
     Note that maximizing (\ref{eqn:termswithw}) with respect to \(\bm{w}\) maximizes (\ref{eqn:derivative}). Let us assume that the index that maximizes the inner maximization with respect to \(m\) is \(\bar{m}\). Therefore, (\ref{eqn:termswithw}) is now rewritten as,
     \begin{align}
         \bigg( \sum_{j=1}^k \sigma(\bm{x}^T \bm{w}^j) - y \bigg)  \nabla^2 \sigma(\bm{x}^T \bm{w}^{\bar{m}}) + \sum_{j=1}^k \big(\nabla \sigma(\bm{x}^T \bm{w}^j) \big)^2 .
     \end{align}
     We use \( a-b \leq |a|+ |b|\) on the first term. Combining the terms corresponding to \(\bar{m}\) and using (\ref{eqn:func2ndder}) to bound the second derivative, 
     \begin{align}
         & \max_{\bm{w}} \bigg[ \frac{1}{10}\big( \sum_{j=1, j \neq \bar{m}}^k \sigma(\bm{x}^T \bm{w}^j) + |y| \big) + \sum_{j=1, j \neq \bar{m}}^k \big(\nabla \sigma(\bm{x}^T \bm{w}^j) \big)^2 \nonumber \\ 
         & + \sigma(\bm{x}^T \bm{w}^{\bar{m}})
         \nabla^2 \sigma(\bm{x}^T \bm{w}^{\bar{m}}) +  \big(\nabla \sigma(\bm{x}^T \bm{w}^{\bar{m}})
          \big)^2 \bigg].
     \end{align}
     We then maximize each of these terms individually leading to the following bounds:
     \begin{itemize}
         \item \(\frac{\sigma(x)}{10} + (\nabla \sigma(x))^2 \leq 0.1176  \quad \forall x\)
         \item \(\sigma(x) \nabla^2 \sigma(x) + (\nabla \sigma(x))^2 \leq 0.0770 \quad \forall x\).
     \end{itemize}
     Incorporating the above, we get the following bound on the gradient Lipschitz constant
     \begin{align} \label{eqn:bound2sig}
         \alpha^* \leq \big[ (k-1) 0.1176 + \frac{|y|}{10} + 0.0770\big] \norm{\bm{x}}^2.
     \end{align}
      Depending on the value of \(k\) and \(y\), we find that either of the bounds in (\ref{eqn:bound1sig}) and (\ref{eqn:bound2sig}) can prove tighter. As both of them are upper bounds, we pick the least one of them.
      The final expression for the upper bound on the gradient Lipschitz constant when a single data point \((\bm{x},y)\) is taken is given by,
      \begin{align}
          \alpha^* &\leq \min(\frac{|k-y|}{10} + \frac{k}{16}, 0.1776(k-1)  \nonumber \\
          & \hspace{2cm}+ \frac{|y|}{10} + 0.0770) \norm{\bm{x}}^2 .
      \end{align}
    \end{proof}
    \noindent
      We now wish to extend this to multiple data points \((\bm{x}(i),y(i))\) for \(i = 1, \cdots N\).
    \revise{When we follow the same derivation for a loss function constructed with multiple data points, the derived upper bound results in the average of the individual upper bounds. This is a direct implication from the fact that 
\(
    \nabla^2 \left( \sum_i f_i \right)= \sum_i \nabla^2 f_i.
\)}

     Therefore, the bound on \(\alpha^*\) is given by 
     \begin{align}
          \alpha^* &\leq \dfrac{1}{N} \sum_{i=1}^N \min\bigg[\frac{|k-y(i)|}{10}+ \frac{k}{16}, \nonumber \\
          & \hspace{1cm}0.1776(k-1)  + \frac{|y(i)|}{10} + 0.0770 \bigg] \norm{\bm{x}(i)}^2. \label{eqn:1lsigfinal}
      \end{align}
      \revise{As the loss function for multiple data points is defined as the average over loss using each of the data points, we note that the derived upper bound on the gradient Lipschitz constant also follows a similar structure.}
      Given the number of neurons in the hidden layer, \(k\), and the data set, the upper bound can be found by simply evaluating the expression derived in (\ref{eqn:1lsigfinal}); the inverse of this bound gives a learning rate which always guarantees that GD will converge. The bound increases with increase in width of the network as well as the norm of the input. It is noted that the bound depends on data only through its norm. Therefore, if different data sets with similar Euclidean norms are encountered, the derived bound can simply be reused.
      \vspace{-0.4cm}
      \subsection{ReLU activation}
The ReLU activation function is given by \(s(x) =\max(0,x)\). Initially, consider a single data point \((\bm{x},y)\) for deriving the gradient Lipschitz constant. 
     
     \begin{theorem}\label{theorem:optConv1LReLU}
     The gradient Lipschitz constant for a single-hidden layer feedforward network with ReLU activation when considering quadratic loss function in (\ref{eqn:loss1L}) when \(act(.) = s(.)\) and \(N=1\) is given by
         \begin{equation}
            \alpha^* = k \norm{\bm{x}}^2.
        \end{equation}
     \end{theorem}
     
     \begin{proof}
     Please see Appendix \ref{Appendix_A}.
     \end{proof}

     We now extend the derivation of the gradient Lipschitz constant for a multiple input database. \cutt{The loss function for the case of \(N\) inputs when the data set \((\bm{x}(i),y(i))\) for \(i = 1,...N\) is employed is given by 
     \begin{equation} \label{eqn:lossNinputsrelu}
              l(\bm{w}) = \dfrac{1}{2N} \sum_{i = 1}^N \left(  \left(\sum_{j=1}^k s(\bm{x}(i)^T \bm{w}^j) \right) - y(i) \right)^2.
     \end{equation}}
     The result in Theorem \ref{theorem:optConv1LReLU} can be extended to \(N\) inputs as
     \begin{equation} \label{eqn:alphaNinputsrelu}
         \alpha^* = \dfrac{1}{N} \max_{\bm{w}} \lambda_{\max}\left( \sum_{i=1}^N  \bm{a(x}(i)\bm{,w}) \bm{a(x}(i)\bm{,w})^T  \right) ,
     \end{equation}
     where
      \begin{equation} \label{eqn:avecmain}
            \bm{a(x,w}) \triangleq \begin{bmatrix} \mathbb{I}_{\{\bm{x}^T\bm{w}^1 \geq 0\}}  \bm{x} & \hdots & \mathbb{I}_{\{\bm{x}^T\bm{w}^k \geq 0\}}  \bm{x} \end{bmatrix}^T.
        \end{equation}
     This involves a maximization over all possible weights and we would like to derive a closed form expression.
     It is observed that the Hessian matrix in this specific problem is structured; it is the sum of outer products of the vector \(\bm{a(x}(i)\bm{,w})\) where the vector consists of \(\bm{x}(i)\) multiplied by appropriate indicators. We wish to exploit the structure of the Hessian matrix to arrive at an elegant solution which can be easily evaluated. Towards that end, we state and prove the following lemma.
     \begin{lemma}  \label{lemma:lem1}
         For a vector \(\bm{a}(\bm{x}(i),\bm{w})\) as defined in (\ref{eqn:avecmain}), the following relation holds
         \begin{align} 
             &\lambda_{max}\left( \sum_{i=1}^N  \bm{\bar{a}(x(i))} \bm{\bar{a}(x(i))} ^T \right) \geq \nonumber \\
             & \qquad\lambda_{max}\left( \sum_{i=1}^N \bm{a(x}(i)\bm{,w}) \bm{a(x}(i)\bm{,w})^T  \right) \quad \forall \bm{w} \label{eqn:upper_alpha}
         \end{align}
         where
         \begin{equation} \label{eqn:abarmain}
            \bm{\bar{a}(x)} \triangleq \begin{bmatrix} \bm{x}  & \hdots & \bm{x} \end{bmatrix}^T \hspace{1cm} \text{(\(k\) terms)}.
        \end{equation}
     \end{lemma}
     \begin{proof}
     Please see Appendix \ref{Appendix_B}.
     \end{proof}
     
    Lemma \ref{lemma:lem1} holds for all values of \(\bm{w}\); therefore, it also holds for that \(\bm{w}\) which maximizes the maximum eigenvalue in (\ref{eqn:alphaNinputsrelu}). In essence, Lemma \ref{lemma:lem1} provides an upper bound on the constant \(\alpha^*\).
     It is also noted that as \(\bm{\bar{a}(x}(i)\bm{)}\) is an instance of \(\bm{a(x}(i)\bm{,w)}\) for a specific \(\bm{w}\),
     \begin{align} 
              &\max_{\bm{w}} \lambda_{max}\left( \sum_{i=1}^N \bm{a(x}(i)\bm{,w}) \bm{a(x}(i)\bm{,w})^T  \right) \geq \nonumber\\
              & \qquad \lambda_{max}\left( \sum_{i=1}^N  \bm{\bar{a}(x}(i)\bm{)} \bm{\bar{a}(x}(i)\bm{)} ^T \right). \label{eqn:lower_alpha}
     \end{align}
     Hence, (\ref{eqn:lower_alpha}) gives a lower bound on the constant \(\alpha^*\).
     \imp{From (\ref{eqn:upper_alpha}) and (\ref{eqn:lower_alpha}), it is evident that the upper and the lower bounds coincide and must be equal to the exact value of \(\alpha^*\), i.e.,}
     \begin{equation} \label{eqn:1LReLUfinal}
         \alpha^* = \lambda_{max} \left( \dfrac{1}{N}  \sum_{i=1}^N  \bm{\bar{a}(x}(i)\bm{)} \bm{\bar{a}(x}(i)\bm{)} ^T \right).
     \end{equation}
     The exact gradient Lipschitz constant for a single hidden layer network with ReLU activation has been derived in (\ref{eqn:1LReLUfinal}). We no longer need the perform the brute force maximization over all weight values as was required in (\ref{eqn:alphaNinputsrelu}). Instead evaluating \(\alpha^*\) is now reduced to finding the maximum eigenvalue of a \(kd \times kd\) matrix. We note that the value of \(\alpha^*\) only depends on the data vectors \(\bm{x}(i)\) and the number of neurons \(k\). As the dimension of the problem increases, the eigenvalue computation will get intensive; in such cases, we can employ well-established bounds like Gershgorin and Brauer's ovals of Cassini to provide easily computable upper bounds on \(\alpha^*\). For convenience, these theorems are stated here.
     
     \begin{theorem}[Gershgorin’s Circles theorem \cite{varga2009matrix}] \label{theorem:gershgorin}
        For a square matrix \(\bm{A}\), the upper bound on the maximum eigenvalue is,
        \begin{equation}
            \lambda_{max}(A) \leq \max_i (a_{ii} + R_i(\bm{A})),
        \end{equation} 
        where \(R_i(\bm{A}) = \sum_{i \neq j} |a_{ij}|\)
     \end{theorem}
     
     \begin{theorem}[Brauer’s Ovals of Cassini] \label{theorem:Brauer}
     For a square matrix \(\bm{A}\), the upper bound on the maximum eigenvalue is given by
        \begin{equation} \label{eqn:alpha2}
          \lambda_{max}(A) \leq \max_{i \neq j} \left(\dfrac{a_{ii} + a_{jj}}{2} + \sqrt{(a_{ii} - a_{jj})^2 + R_i(\bm{A}) R_j(\bm{A})}  \right) ,
     \end{equation}
     where \(R_i(\bm{A}) = \sum_{i \neq j} |a_{ij}|\).
     \end{theorem}
     The bound in Theorem \ref{theorem:Brauer} is guaranteed to be provide a bound which is not worse than the Gershgorin bound \cite{deville2019optimizing}.
     The bounds stated above can be used to provide an upper bound on the gradient Lipschitz constant if eigenvalue computation is a constraint. The inverse of the derived constant \(\alpha^*\) or its upper bound can be used as the learning rate while training the network, and this will guarantee convergence of GD. 

    \section{Deriving the gradient Lipschitz constant for a two hidden layer neural network} \label{sec:2layer}

 Here, we focus on a shallow architecture with two hidden layers between the input and output layers as illustrated in Fig. \ref{fig:2Larch}. The weight matrix between the input and the first hidden layer is denoted as \(\bm{V} \in \mathbb{R}^{d \times k_1}\) where \(k_1\) is the number of neurons in the first hidden layer. The weight matrix between the two hidden layers is denoted as \(\bm{W} \in \mathbb{R}^{k_1 \times k_2}\) where \(k_2\) is the number of neurons in the second hidden layer. The output of the network is the sum of the outputs of the neurons in the second hidden layer. Let us denote the parameters of the network \(\bm{V},\bm{W}\) as a single vector \(\bm{\theta}\). Note that the dimension of \(\bm{\theta}\) is \(k_1(d+k_2)\).
     \begin{equation} \label{eqn:theta}
         \bm{\theta} = \begin{bmatrix} 
                        {\bm{V^1}}^T \hdots {\bm{V^{k_1}}}^T & {\bm{W^1}}^T \hdots {\bm{W^{k_2}}}^T
                  \end{bmatrix}^T
     \end{equation}

\tikzset{%
  every neuron/.style={
    circle,
    draw,
    minimum size=0.8cm
  },
  neuron missing/.style={
    draw=none, 
    scale=4,
    text height=0.333cm,
    execute at begin node=\color{black}$\vdots$
  },
}
\begin{figure}
\centering
\begin{tikzpicture}[scale = 0.7,x=1.5cm, y=1.5cm, >=stealth]

\foreach \m/\l [count=\y] in {1,2,3,missing,4}
  \node [every neuron/.try, neuron \m/.try] (input-\m) at (0,2.5-\y) {};

\foreach \m [count=\y] in {1,missing,2}
  \node [every neuron/.try, neuron \m/.try ] (hidden1-\m) at (1.33,2-\y*1.25) {};

\foreach \m [count=\y] in {1,missing,2}
  \node [every neuron/.try, neuron \m/.try ] (hidden2-\m) at (2.67,2-\y*1.25) {};
  
\foreach \m [count=\y] in {1}
  \node [every neuron/.try, neuron \m/.try ] (output-\m) at (4,-0.2) {$+$};

\foreach \l [count=\i] in {1,2,3,n}
  \draw [<-] (input-\i) -- ++(-1,0)
    node [above, midway] {$x_\l$};

\foreach \l [count=\i] in {1}
  \node [above] at (hidden1-\i.north) {$act(.)$};
  
\foreach \l [count=\i] in {2}
  \node [below] at (hidden1-\l.south) {$act(.)$};  
  
  \foreach \l [count=\i] in {1}
  \node [above] at (hidden2-\i.north) {$act(.)$};
  
   \foreach \l [count=\i] in {2}
  \node [below] at (hidden2-\l.south) {$act(.)$};

\foreach \l [count=\i] in {1}
  \draw [->] (output-\i) -- ++(1,0)
    node [above, midway] {$y$};

\foreach \i in {1,...,4}
  \foreach \j in {1,...,2}
    \draw [->] (input-\i) -- (hidden1-\j);
    
\foreach \i in {1,...,2}
  \foreach \j in {1,...,2}
    \draw [->] (hidden1-\i) -- (hidden2-\j);

\foreach \i in {1,...,2}
  \foreach \j in {1}
    \draw [->] (hidden2-\i) -- (output-\j);

\foreach \l [count=\z from 0] in {Input, Hidden, Hidden, Output}
  \node [align=center, above] at (\z*1.3,2) {\l \\ layer};

\end{tikzpicture}
   \caption{The architecture of a two hidden layer network}
    \label{fig:2Larch}
\end{figure}
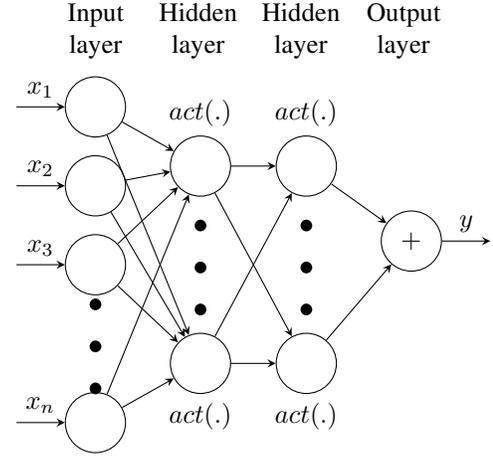  
   This derivation is challenging as it is not a straight-forward extension of the single layer case; the Hessian involves two weight matrices \(\bm{V},\bm{W}\) to be optimized over. The squared loss function is considered whose expression is given by,
\begin{align}
         l(\bm{\theta}) = \dfrac{1}{2N} \sum_{i=1}^N \Bigg( \bigg( \sum_{l_2 = 1}^{k_2} &
         act \big(\sum_{l_1 = 1}^{k_1} act (\bm{x}(i)^T  \bm{V^{l_1}}) \nonumber\\
         & W_{l_1 l_2} \big) \bigg)- y(i) \Bigg)^2. \label{eqn:loss2L}
     \end{align}

\subsection{Sigmoid activation}
Here, we derive the gradient Lipschitz constant of a 2-hidden layer network with sigmoid activation function. As done previously, we initially consider a single data tuple \((\bm{x},y)\) where \(\bm{x} \in \mathbb{R}^d\) and \(y \in \mathbb{R}\). 

     \begin{theorem} \label{theorem:2layerSigConv}
        The gradient Lipschitz constant for a two hidden layer feedforward network with sigmoid activation when considering quadratic loss function in (\ref{eqn:loss2L}) with \(act(.)=\sigma(.)\) and \(N=1\) is given by,
         \begin{align}
          \alpha^* & \leq k_1\left(\frac{k_2 \beta \norm{\bm{x}}}{16} \right)^2 + \frac{k_1 k_2}{16} +\max\bigg( \frac{1}{10} +  \nonumber\\
          & \left[\frac{1}{4} + \frac{\beta}{10}\right] \frac{ k_1 \norm{\bm{x}}_1}{4},
          \left[\frac{1}{4} + \frac{\beta}{10}\right] \frac{k_2 \norm{\bm{x}}_{\infty}}{4} + \nonumber\\
         & \left[\frac{\beta}{1000}  + \frac{1}{4}\right] k_1 k_2 \beta \norm{\bm{x}}_1 \norm{\bm{x}}_{\infty} \bigg)  |k_2 -y|
     \end{align}
     when \(|\theta_i|< \beta \quad \forall i\) for \(\bm{\theta}\) as defined in (\ref{eqn:theta}).
      \end{theorem}
      \begin{proof}
        Please see Appendix \ref{Appendix_C}.
    \end{proof}    
    This is further extended to the case of \(N\) inputs and the obtained constant is given by 
     \begin{align}
         \alpha^* &\leq \dfrac{1}{N}\sum_{i=1}^N \bigg[ k_1\left(\frac{k_2 \beta \norm{\bm{x}(i)}}{16} \right)^2 + \frac{k_1 k_2}{16} +  \max\bigg( \frac{1}{10} +   \nonumber \\
         & \left[\frac{1}{4} + \frac{\beta}{10}\right] \frac{ k_1 \norm{\bm{x}(i)}_1}{4}, \left[\frac{1}{4} + \frac{\beta}{10}\right] \frac{k_2 \norm{\bm{x}(i)}_{\infty}}{4} +\nonumber\\
         & \left[\frac{\beta}{1000}  + \frac{1}{4}\right]  k_1 k_2 \beta \norm{\bm{x}(i)}_1 \norm{\bm{x}(i)}_{\infty} \bigg)  |k_2 -y(i)|\bigg].
     \end{align}
We note that increase in dimension of the architecture will lead to an increase in the bound. The derived bound also depends on the maximum value in the weight matrix. Therefore, the bound is tighter when there are no spurious values with large magnitude in the weight matrix. 

\vspace{-0.4cm}
\subsection{ReLU Activation} 
Initially, consider a single data tuple \((\bm{x},y)\) where \(\bm{x} \in \mathbb{R}^d\) and \(y \in \mathbb{R}\). 
     
     \begin{theorem}\label{theorem:optConv2LReLU}
         The gradient Lipschitz constant for a two hidden layer feedforward network with ReLU activation when considering quadratic loss function in (\ref{eqn:loss2L}) with \(act(.) = s(.)\) and \(N=1\) is given by
            \begin{equation}
         \alpha^* \leq k_1(d+k_2) \beta^2 \norm{\bm{x}}^2 + \max( A_{max} k_2 |\bm{x}|_{\infty}, A_{max} |\bm{x}|_1),
     \end{equation}
     where \(A_{max} = k_1 k_2 \beta^2 \norm{\bm{x}} -y\) when \(|\theta_i|< \beta \quad \forall i\).
      \end{theorem}
      \begin{proof}
        Please see Appendix D.
     \end{proof}
     
     
    Extending to a database of \(N\) inputs, i.e., \((\bm{x}(i),y(i))\) for \(i=1,...N\), the following bound is obtained on the gradient Lipschitz constant,
    
    \begin{align} \label{eqn:final2lrelu}
         \alpha^* &\leq \dfrac{1}{N} \sum_{i=1}^N \bigg(k_1(d+k_2) \beta^2 \norm{\bm{x}(i)}^2 + \nonumber \\
         & \quad \quad \max( (A_{max}(i) k_2 |\bm{x}(i)|_{\infty}, A_{max}(i) |\bm{x}(i)|_1)  \bigg),
     \end{align}
     where \(A_{max}(i) = k_1 k_2 \beta^2 \norm{\bm{x}(i)} -y(i).\) The derived bound depends on the dimension of the problem and on the factor \(\beta\) which is the maximum magnitude in the weight matrix. The bound increases linearly with increase in any of the following parameters: \(k_1,k_2,d\) and quadratically on \(\beta\). 
    \section{Proposed search algorithm} \label{sec:proposed}
We propose an algorithm that uses the derived bounds from previous sections to arrive at a learning rate which exhibits faster convergence than using the inverse of the derived bound. 

\vspace{-0.4cm}
\subsection{Why is a search algorithm required?}
\cut{Let the exact gradient Lipschitz constant of the objective be denoted as \(\alpha^*\) and the derived upper bound as \(\alpha\); from Section \ref{sec:motiv}, we recall that the optimal learning rate is \(\eta^* = 1/ \alpha^*\), i.e., \(\eta^*\) is the learning rate that results in the fastest convergence over all possible initialization of weights.} For a derived upper bound \(\alpha\), the corresponding learning rate is found as \(\eta = 1/\alpha\). Note that \(\eta<\eta^*\) (where \(\eta^*=1/\alpha^*\)) and therefore, any learning rate derived from an upper bound is guaranteed to result in non-increasing traces for GD. However, there may exist learning rates that are greater than \(\eta\) which lead to faster convergence. 

Even when the exact value of gradient Lipschitz constant is available, optimality over all possible initializations is considered. However, in practical scenarios, the range of values with which the neural networks are initialized are restricted and hence, we do not require a universally optimal learning rate. In other words, we can afford to have learning rates even higher than \(\eta^*\) as long as it guarantees monotonically decreasing iterates in the region where the weights are initialized. 

Summarizing, the motivations for proposing a search are two-fold: the derived bounds may be loose which gives room for finding better learning rates, and we wish to exploit the weight initialization to find a learning rate customized to the initialization.

\vspace{-0.4cm}
\subsection{Proposed algorithm}
The search is for a learning rate which leads to faster convergence than the inverse of the derived bound, while ensuring that it produces decreasing iterates. This search can be conducted by employing a search interval customized for a given data set and weight initialization. \added{The start-of-the-art hyper-parameter tuning libraries such as HyperOpt\cite{bergstra2013making} allow the user to set the search space.} In our work, we adopt the HyperOpt\footnote{In this manuscript, we refer to the TPE implementation in the HyperOpt library as simply HyperOpt.} implementation of TPE \cite{bergstra2013making}. As they do not utilize the information regarding the objective, the search for learning rate is typically conducted in the interval \([0,1]\).\\
\begin{algorithm}  
    \caption{Binary search algorithm}    \label{alg:BinSearch}
    \begin{algorithmic}[1]
        \State \textbf{Input}: Derived bound \(\alpha\), Evaluations \(E\), Epochs \(T\)
        \State \textbf{Initialization:} \(\eta = 1/\alpha\), \(l_c = \alpha\), \(loss^* = Loss(\eta,T)\)
        \For { \(i = 1,2,\cdots E\)} 
            \State Run GD for and observe \(Loss(\eta,T)\).
            \If{\(Loss(\eta) < loss^*\) AND Iterates are non-increasing}
                \State \(loss^*=Loss(\eta,T)\) \hspace{1.5cm} \text{(Update best loss)}
                \State \(l_c = \eta^{-1}\)
                \State \(\eta = (l_c/2)^{-1}\) \hspace{1.5cm} \text{(Increase learning rate)}
            \Else 
                \State \(\eta = \left[\dfrac{\eta^{-1} + l_c}{2}\right]^{-1}\) \hspace{0.7cm} \text{(Decrease learning rate)}
            \EndIf
        \EndFor
        \State \textbf{Output:} Learning rate = \(\eta\)
    \end{algorithmic}
\end{algorithm}
The algorithm is inspired from binary search \cite{davis1969binary}. The algorithm is initialized with a learning rate that is guaranteed to converge (i.e., \(1/\alpha\) where \(\alpha\) is the derived bound) and is allowed a certain number of trials. If the learning rate chosen in a trial results in a converging trace of GD, a higher learning rate is chosen for the next trial; else, a lower learning rate is chosen. The learning rate leading to the lowest loss is reported at the end of the search algorithm. We note that as the algorithm is initialized with a convergent learning rate, it never yields a divergent learning rate, unlike other search algorithms like grid search, random search and HyperOpt.\\
\revise{Note that one can apply more sophisticated search techniques to carry out this search; we adopt the BinarySearch algorithm as it is intuitive and effective. When the ends of a search interval are known, binary search is typically employed in many applications such as \cite{biernacki2016model}. In our implementation, the BinarySearch algorithm checks if the midpoint of the interval results in a learning rate that gives monotonically decreasing iterates. If so, it searches through the lower interval, else, it chooses the higher interval.\\}
The proposed algorithm is described in Algorithm \ref{alg:BinSearch}. Note that in the algorithm, \(Loss(\eta,T)\) refers to the value of the loss function at the end of \(T\) epochs using the learning rate \(\eta\). 


\vspace{-0.4cm}
\subsection{Advantages and remarks}
The inverse of the derived gradient Lipschitz constant always acts as a valid learning rate. Therefore, in applications where a slower convergence is acceptable, this method is highly useful since it allows one to actually skip hyperparameter tuning altogether.

In other search methods, the search space is often considered as \([0,1]\). However, there may be applications where the inverse of the gradient Lipschitz constant is greater than one. This in turn implies that the proposed method will choose learning rates greater than one whilst guaranteeing convergence whereas the traditional methods with restricted search space, say \([0,1]\) will choose a learning rate less than 1. 

In the case that the optimal learning rate is of a very low order, search algorithms like random search or HyperOpt may always encounter diverging behaviour even after the allotted number of evaluations are utilized. However, in the case of the proposed BinarySearch algorithm, we are guaranteed to find a learning rate which would result in a successful GD epoch.
These advantages are demonstrated with the help of simulations in the forthcoming section.

\section{Simulation Results} \label{sec:sim}
The effectiveness of the proposed algorithm is compared against HyperOpt. As HyperOpt is already shown to outperform random search \cite{bergstra2013making}, we only compare with the HyperOpt tool that uses the TPE. To do so, we run \(100\) experiments with the same number of evaluations allotted for both HyperOpt and BinarySearch. To compare optimization strategies, we can opt for any of the following metrics\footnote{\url{https://sigopt.com/blog/evaluating-hyperparameter-optimization-strategies/}}:
\begin{itemize}
    \item Best-found value: The loss achieved during the best-performing evaluation in an experiment is compared and the fraction of times BinarySearch outperforms HyperOpt is tabulated.
    \item Best trace: The best trace for both the competing algorithms are compared. The learning rate leading to the least area under the convergence curve is said to yield the best trace.
\end{itemize}
\revise{Note that using best loss as the only metric for comparing two optimization techniques may not be sufficient. For example, consider two optimization mechanisms that reach the same minimum in 100 and 1000 epochs. The best loss metric ranks both algorithms equally whereas convergence in 100 epochs is preferable. As the best trace metric compares the area below the convergence curves, it ranks the algorithm using 100 epochs higher than the other. The speed of convergence especially becomes important while training complex models which take significant amount of time to train.}
The synthetic simulation is inspired from the setting in works like \cite{zhang2019learning,li2017convergence} that deal with the theoretical properties of shallow networks. We consider a database with points \((\bm{x}(i),y(i))\) for \(i=1,..N\) where \(\bm{x}(i) \sim \mathcal{N}(\bm{0},I)\) similar to \cite{zhang2019learning}. It is assumed that there is an underlying network known as the teacher network with weights \(\bm{w^*}\). The weights of the teacher network are also sampled from a zero mean unit variance Gaussian distribution. The corresponding labels \(y(i)\) are generated by passing the data through the teacher network. For our simulations, we consider \(N=100\) with \(T\) epochs.

The network to be trained is referred to as the student network. The weights of the student network are initialized using Xavier initialization \cite{glorot2010understanding} and the quadratic loss function is employed. The optimization algorithm used for training is GD and it is run for \(T\) epochs. The algorithms, both BinarySearch and HyperOpt, are allowed a fixed number of evaluations. This is repeated for 100 experiments (each with a different database and weight initialization). All results reported are over 100 experiments.

\vspace{-0.4cm}
\subsection{One hidden layer networks}
\subsubsection{Comparison with HyperOpt}
For a single hidden layer, we run GD for \(T=100\) epochs. We note that the best learning rate chosen by HyperOpt after the stipulated number of evaluations sometimes still lead to unsuccessful GD epochs in case of ReLU activation, i.e., the iterates diverge while our method never leads to divergent behaviour. The fraction of times that divergent behaviour is observed for HyperOpt is tabulated in Table \ref{tab:BinvsHyper1LReluDiv}. In the remaining successful experiments, we compare the final loss obtained using the learning rate chosen by both BinarySearch and HyperOpt. The fraction of experiments in which BinarySearch outperforms (results in a lower 'best-found value' than) HyperOpt is tabulated in Table \ref{tab:BinvsHyper1L}. 
\begin{table}[h]
    \centering
    \begin{tabular}{|c|c|c|c|c|}
        \hline
         {} & {} & \multicolumn{3}{c|}{No. of evaluations}\\
        \hline
         d & k & 5 & 10 & 20 \\
         \hline
         10 & 10  & 0.12 & 0.01 & 0 \\
         20 & 5  & 0.03 & 0 &  0  \\
         5 & 20 & 0.18 & 0.07 & 0\\
         20 & 20  & 0.37 & 0.07 & 0.02\\
         \hline
    \end{tabular}
    \caption{Fraction of times HyperOpt diverges for 1 hidden layer network with ReLU activation}
    \label{tab:BinvsHyper1LReluDiv}
\end{table}

\begin{table}[h]
    \centering
    \begin{tabular}{|c|c|c|c|c|c|c|c|}
        \hline
        {} & {} & \multicolumn{3}{c|}{ReLU activation} & \multicolumn{3}{c|}{Sigmoid activation}\\
        \hline
        {} & {} & \multicolumn{3}{c|}{No. of evaluations} & \multicolumn{3}{c|}{No. of evaluations}\\
        \hline
         d & k & 5 & 10 & 20 & 5 & 10 & 20 \\
         \hline
         10 & 10 & 0.81 & 0.91 & 0.93 & 1 &   1 &   1\\
         20 & 5 & 0.76 &  0.85 & 0.85 & 1 &   1 &   1 \\
         5 & 20 & 0.74 & 0.90 & 0.95 &  1 &   1 &   1\\
         20 & 20 &  0.73 & 0.86 & 0.89&  1 &   1 &   1 \\
         \hline
    \end{tabular}
    \caption[Caption for LOF]{Fraction of times the best value for BinarySearch outperforms HyperOpt for 1 hidden layer network out of successful experiments\protect\footnotemark}
    \label{tab:BinvsHyper1L}
\end{table}
\footnotetext{Experiments in which HyperOpt diverges are not considered.}

We notice that for higher number of evaluations, BinarySearch always outperforms HyperOpt.
\added{It should be noted that these comparisons are performed after eliminating the experiments for which HyperOpt diverges. For instance, for the configuration \(d=k=20\) with 5 evaluations using ReLU activation, BinarySearch outperforms HyperOpt \(73\%\) out of the \(100-37=63\) successful experiments. If we also consider the divergent experiments, BinarySearch outperforms HyperOpt \(83\%\) of the times.} In the case of sigmoid activation, the divergent behaviour is not observed. As the gradient of the loss function is of a small order of magnitude (in the order of \(10^{-2}\)), GD does not diverge for higher learning rates. Also, the learning rate derived as the inverse of gradient Lipschitz constant for \(d=2,k=3,N=100\) is \(1.78\) which itself is greater than \(1\). This implies that any learning rate less than \(1.78\) will never lead to divergent behaviour and learning rates greater than \(1.78\) can be explored. \added{One can argue that the derived bound can be used to modify the search interval of existing algorithms; this is discussed at the end of this section.}

\begin{figure}[h]
    \centering
    \resizebox{\linewidth}{!}{\begin{tikzpicture}[thick]
    \begin{axis}[
        width=6cm,
        height=4.5cm,
        xmin=0,
        xmax=80,
        ymin=0.00,
        ymax=15,
        grid=major,
        xlabel={Time steps},
        ylabel={Loss \(l(\theta)\)},
        xlabel style={at={(0.50,0.05)}},
        ylabel style={at={(0.06,0.50)}},
        xtick={0.00,20.00,...,80.00},
        ytick={0.00,5.00,...,20.00},
        log ticks with fixed point,
        legend pos=north east,
        legend cell align={left},
        legend style={fill opacity=0.6, draw opacity=1.0, text opacity=1.0, font=\small}
        ]
        
        \pgfplotsset{every tick label/.append style={font=\small}}
        
        \addplot[red, solid, thick, line width = 1pt, mark=triangle, mark size={1.5}, mark repeat=10, mark phase=10] 
            table [x=x_data, y=y_BinarySearch, col sep=comma]{./ReLU_L1_e10.csv};
        \addlegendentry{BinarySearch};
        
        \addplot[blue, solid, thick, line width = 1pt, mark=square, mark size={1.5}, mark repeat=10, mark phase=10] 
            table [x=x_data, y=y_HyperOpt, col sep=comma]{./ReLU_L1_e10.csv};
        \addlegendentry{HyperOpt};

    \end{axis}
\end{tikzpicture}}
    \caption{Best trace comparison of single hidden layer ReLU network with \(d=k=10\) with 10 evaluations}
    \label{fig:Relu_1L}
\end{figure}
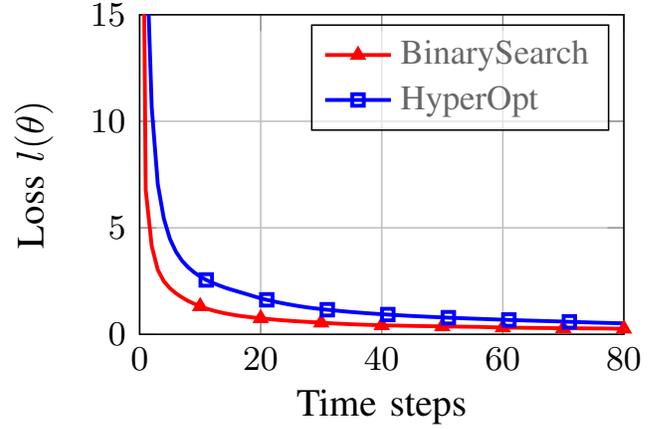

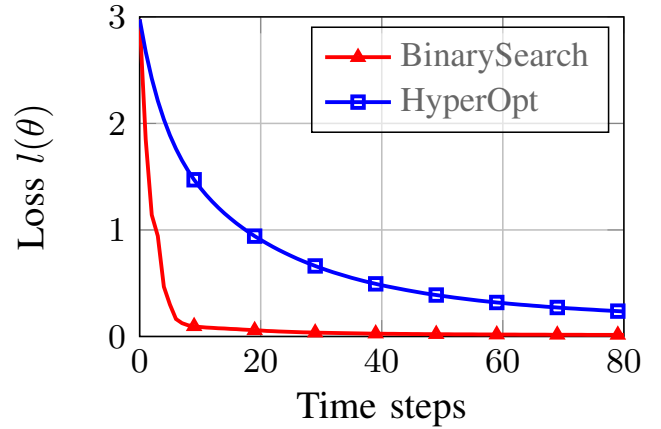
\begin{figure}[h]
    \centering
    \resizebox{\linewidth}{!}{\begin{tikzpicture}[thick]
    \begin{axis}[
        width=6cm,
        height=4.5cm,
        xmin=0,
        xmax=80,
        ymin=0.00,
        ymax=3,
        grid=major,
        xlabel={Time steps},
        ylabel={Loss \(l(\theta)\)},
        xlabel style={at={(0.50,0.05)}},
        ylabel style={at={(0.06,0.50)}},
        xtick={0.00,20.00,...,80.00},
        ytick={0.00, 1.00,...,3.00},
        log ticks with fixed point,
        legend pos=north east,
        legend cell align={left},
        legend style={fill opacity=0.6, draw opacity=1.0, text opacity=1.0, font=\small}
        ]
        \pgfplotsset{every tick label/.append style={font=\small}}
        
        \addplot[red, solid, thick, line width = 1pt, mark=triangle, mark size={1.5}, mark repeat=10, mark phase=10] 
            table [x=x_data, y=y_BinarySearch, col sep=comma]{./sig_L1_e10.csv};
        \addlegendentry{BinarySearch};
        
        \addplot[blue, solid, thick, line width = 1pt, mark=square, mark size={1.5}, mark repeat=10, mark phase=10] 
            table [x=x_data, y=y_HyperOpt, col sep=comma]{./sig_L1_e10.csv};
        \addlegendentry{HyperOpt};

    \end{axis}
\end{tikzpicture}}
    \caption{Best trace comparison of single hidden layer sigmoid network with \(d=k=10\) with 10 evaluations}
    \label{fig:sig_1L}
\end{figure}

The metric, 'best-found value' grades an algorithm based on the final loss value that the algorithm converges to. We also need a metric to quantify the performance in terms of the convergence rate. Hence, we also provide the best-trace metric, where the tuning strategies are compared based on their convergence. The curve with the fastest convergence (least area under convergence curve) out of all 100 experiments is plotted for BinarySearch and HyperOpt. We provide the results for a specific configuration with \(d=k=10\) where each method is allowed 10 evaluations for ReLU and sigmoid activation functions in Fig. \ref{fig:Relu_1L} and \ref{fig:sig_1L} respectively. We note that the proposed method results in better convergence curves than the existing method, HyperOpt.

\revise{In order to further study the attributes of the learning rates chosen by the proposed method in comparison with HyperOpt, we tabulate the mean, standard deviation, maximum and minimum values chosen over the 100 experiments. We study this for the case of \(d=k=10\) in Table \ref{tab:comp1}.}
\begin{table}[h]
    \centering
   \begin{tabular}{|c|c|c|c|c|c|c|}
    \hline
       Activation & Evals & Algorithm &  Mean & Std. dev & Max & Min\\
    \hline     
     \multirow{4}{*}{ReLU} & \multirow{2}{*}{5} & BinarySearch & 0.324 & 0.033 & 0.409 &  0.227 \\
     & & HyperOpt & 0.234 & 0.095  & 0.459  & 0.017 \\ 
     \cline{2-7}
     & \multirow{2}{*}{20} & BinarySearch & 0.385 & 0.045 & 0.498  & 0.292 \\
     & & HyperOpt & 0.328 & 0.069 & 0.462  & 0.098\\ 
    \hline
    \multirow{4}{*}{Sigmoid} & \multirow{2}{*}{5} & BinarySearch & 3.175 & 0.071 & 3.327 & 2.983\\
     & & HyperOpt & 0.778 & 0.174 & 0.996 & 0.342 \\ 
     \cline{2-7}
     & \multirow{2}{*}{20} & BinarySearch & 13.359 & 3.984 & 30.805 & 7.514\\
     & & HyperOpt & 0.868 & 0.095 & 0.996 & 0.565 \\ 
    \hline
    \end{tabular}
    \caption{Variation of the chosen learning rates for single hidden layer network}
    \label{tab:comp1}
\end{table}

\revise{Although the above tabulation is for 100 experiments, note that HyperOpt returns a learning rate that results in diverging traces for a small fraction of experiments (0.01); these entries are ignored while computing the tabulated constants for HyperOpt. \\
From Table \ref{tab:comp1}, it is noted that BinarySearch always chooses a larger learning rate on an average as compared to HyperOpt which leads to better convergence. The maximum learning rate (the learning rate yielding the best trace graph) chosen by BinarySearch is greater than that of HyperOpt as the number of evaluations increase as evidenced by the numbers corresponding to ReLU activation. In the case of ReLU activation, it is observed that the proposed method has lesser variance in choosing a step size as compared to HyperOpt. For the sigmoid activation, our method chooses rates much greater than one, as it is allowed by the structure of the problem whereas HyperOpt typically is restricted to the interval \([0,1]\). This also explains why BinarySearch results in a much faster convergence than HyperOpt in this case.
}
\subsubsection{Comparison with other optimization algorithms}
\revise{We also compare the performance of our proposed tuning method against popular optimization algorithms such as Adam \cite{kingma2014adam}, Adagrad \cite{duchi2011adaptive}, Adadelta \cite{zeiler2012adadelta} and RMSProp \cite{hinton2012rmsprop}. Although a default learning rate of 0.001 is suggested for Adam, RMSProp and Adadelta, we note that a learning rate of 0.01 fares better in this case. To demonstrate, we compare against the default learning rate (0.001) as well as a learning rate of 0.01 for the Adam optimizer. For the other optimizers, we only show results for a learning rate of 0.01 which fares better than their default rate of 0.001. The learning curve corresponding to the plain vanilla gradient descent using the derived bound is also included for the comparison.}

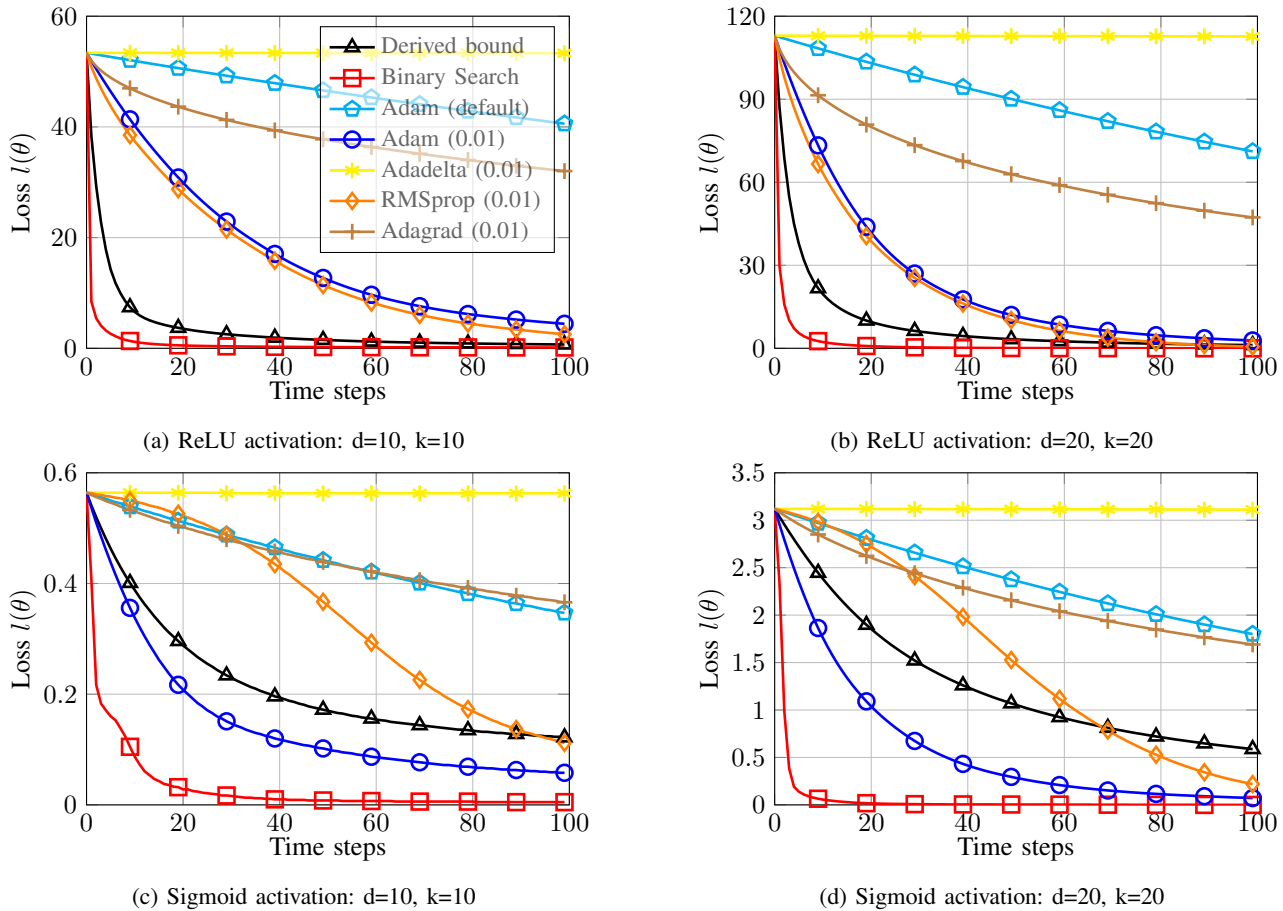
\begin{figure*}[h]
\begin{subfigure}{.5\textwidth}
  \centering
  \begin{tikzpicture}[thick]
    \begin{axis}[
        width=8cm,
        height=6cm,
        xmin=0,
        xmax=100,
        ymin=0.00,
        ymax=60,
        grid=major,
        xlabel={Time steps},
        ylabel={Loss \(l(\theta)\)},
        xlabel style={at={(0.50,0.05)}},
        ylabel style={at={(0.06,0.50)}},
        xtick={0.00,20.00,...,100.00},
        ytick={0.00,20.00,...,60.00},
        log ticks with fixed point,
        legend pos=north east,
        legend cell align={left},
        legend style={fill opacity=0.6, draw opacity=1.0, text opacity=1.0, font=\small}
        ]

        \addplot[black, solid, thick, line width = 1pt, mark=triangle, mark size={3.0}, mark repeat=10, mark phase=10] 
            table [x=x_data, y=y_GD, col sep=comma]{./ReluL1_d10k10.csv};
        \addlegendentry{Derived bound};
        
        \addplot[red, solid, thick, line width = 1pt, mark=square, mark size={3.0}, mark repeat=10, mark phase=10] 
            table [x=x_data, y=y_BinSearch, col sep=comma]{./ReluL1_d10k10.csv};
        \addlegendentry{Binary Search};
        
        \addplot[cyan, solid, thick, line width = 1pt, mark=pentagon, mark size={3.0}, mark repeat=10, mark phase=10] 
            table [x=x_data, y=y_AdamDef, col sep=comma]{./ReluL1_d10k10.csv};
        \addlegendentry{Adam (default)};
        
        \addplot[blue, solid, thick, line width = 1pt, mark=o, mark size={3.0}, mark repeat=10, mark phase=10] 
            table [x=x_data, y=y_AdamNotDef, col sep=comma]{./ReluL1_d10k10.csv};
        \addlegendentry{Adam (0.01)};
        
        
        \addplot[yellow, solid, thick, line width = 1pt, mark=asterisk, mark size={3.0}, mark repeat=10, mark phase=10] 
            table [x=x_data, y=y_AdaDeltaNotDef, col sep=comma]{./ReluL1_d10k10.csv};
        \addlegendentry{Adadelta (0.01)};
        
        \addplot[orange, solid, thick, line width = 1pt, mark=diamond, mark size={3.0}, mark repeat=10, mark phase=10] 
            table [x=x_data, y=y_RMSprop, col sep=comma]{./ReluL1_d10k10.csv};
        \addlegendentry{RMSprop (0.01)};
        
        \addplot[brown, solid, thick, line width = 1pt, mark=+, mark size={3.0}, mark repeat=10, mark phase=10] 
            table [x=x_data, y=y_Adagrad, col sep=comma]{./ReluL1_d10k10.csv};
        \addlegendentry{Adagrad (0.01)};





    \end{axis}
\end{tikzpicture} 
  \caption{ReLU activation: d=10, k=10}
  \label{fig:relud10k10}
\end{subfigure}
\begin{subfigure}{.5\textwidth}
  \centering
  \begin{tikzpicture}[thick]
    \begin{axis}[
        width=8cm,
        height=6cm,
        xmin=0,
        xmax=100,
        ymin=0.00,
        ymax=120,
        grid=major,
        xlabel={Time steps},
        ylabel={Loss \(l(\theta)\)},
        xlabel style={at={(0.50,0.05)}},
        ylabel style={at={(0.06,0.50)}},
        xtick={0.00,20.00,...,100.00},
        ytick={0.00,30.00,...,120.00},
        log ticks with fixed point,
        legend pos=north east,
        legend cell align={left},
        legend style={fill opacity=0.6, draw opacity=1.0, text opacity=1.0, font=\small}
        ]

        \addplot[black, solid, thick, line width = 1pt, mark=triangle, mark size={3.0}, mark repeat=10, mark phase=10] 
            table [x=x_data, y=y_GD, col sep=comma]{./ReluL1_d20k20.csv};
        
        \addplot[red, solid, thick, line width = 1pt, mark=square, mark size={3.0}, mark repeat=10, mark phase=10] 
            table [x=x_data, y=y_BinSearch, col sep=comma]{./ReluL1_d20k20.csv};
        
        \addplot[cyan, solid, thick, line width = 1pt, mark=pentagon, mark size={3.0}, mark repeat=10, mark phase=10] 
            table [x=x_data, y=y_AdamDef, col sep=comma]{./ReluL1_d20k20.csv};
        
        \addplot[blue, solid, thick, line width = 1pt, mark=o, mark size={3.0}, mark repeat=10, mark phase=10] 
            table [x=x_data, y=y_AdamNotDef, col sep=comma]{./ReluL1_d20k20.csv};
        
        
        \addplot[yellow, solid, thick, line width = 1pt, mark=asterisk, mark size={3.0}, mark repeat=10, mark phase=10] 
            table [x=x_data, y=y_AdaDeltaNotDef, col sep=comma]{./ReluL1_d20k20.csv};
        
        \addplot[orange, solid, thick, line width = 1pt, mark=diamond, mark size={3.0}, mark repeat=10, mark phase=10] 
            table [x=x_data, y=y_RMSprop, col sep=comma]{./ReluL1_d20k20.csv};
        
        \addplot[brown, solid, thick, line width = 1pt, mark=+, mark size={3.0}, mark repeat=10, mark phase=10] 
            table [x=x_data, y=y_Adagrad, col sep=comma]{./ReluL1_d20k20.csv};





    \end{axis}
\end{tikzpicture}   
  \caption{ReLU activation: d=20, k=20}
  \label{fig:relud20k20}
\end{subfigure}
\\
\begin{subfigure}{.5\textwidth}
  \centering
  \begin{tikzpicture}[thick]
    \begin{axis}[
        width=8cm,
        height=6cm,
        xmin=0,
        xmax=100,
        ymin=0.00,
        ymax=0.6,
        grid=major,
        xlabel={Time steps},
        ylabel={Loss \(l(\theta)\)},
        xlabel style={at={(0.50,0.05)}},
        ylabel style={at={(0.06,0.50)}},
        xtick={0.00,20.00,...,100.00},
        ytick={0.00,0.2,...,0.6},
        log ticks with fixed point,
        legend pos=north east,
        legend cell align={left},
        legend style={fill opacity=0.6, draw opacity=1.0, text opacity=1.0, font=\small}
        ]

        \addplot[black, solid, thick, line width = 1pt, mark=triangle, mark size={3.0}, mark repeat=10, mark phase=10] 
            table [x=x_data, y=y_GD, col sep=comma]{./SigL1_d10k10.csv};
        
        \addplot[red, solid, thick, line width = 1pt, mark=square, mark size={3.0}, mark repeat=10, mark phase=10] 
            table [x=x_data, y=y_BinSearch, col sep=comma]{./SigL1_d10k10.csv};
        
        \addplot[cyan, solid, thick, line width = 1pt, mark=pentagon, mark size={3.0}, mark repeat=10, mark phase=10] 
            table [x=x_data, y=y_AdamDef, col sep=comma]{./SigL1_d10k10.csv};
        
        \addplot[blue, solid, thick, line width = 1pt, mark=o, mark size={3.0}, mark repeat=10, mark phase=10] 
            table [x=x_data, y=y_AdamNotDef, col sep=comma]{./SigL1_d10k10.csv};
        
        
        \addplot[yellow, solid, thick, line width = 1pt, mark=asterisk, mark size={3.0}, mark repeat=10, mark phase=10] 
            table [x=x_data, y=y_AdaDeltaNotDef, col sep=comma]{./SigL1_d10k10.csv};
        
        \addplot[orange, solid, thick, line width = 1pt, mark=diamond, mark size={3.0}, mark repeat=10, mark phase=10] 
            table [x=x_data, y=y_RMSprop, col sep=comma]{./SigL1_d10k10.csv};
        
        \addplot[brown, solid, thick, line width = 1pt, mark=+, mark size={3.0}, mark repeat=10, mark phase=10] 
            table [x=x_data, y=y_Adagrad, col sep=comma]{./SigL1_d10k10.csv};





    \end{axis}
\end{tikzpicture}  
  \caption{Sigmoid activation: d=10, k=10}
  \label{fig:sigd10k10}
\end{subfigure}
\begin{subfigure}{.5\textwidth}
  \centering
  \begin{tikzpicture}[thick]
    \begin{axis}[
        width=8cm,
        height=6cm,
        xmin=0,
        xmax=100,
        ymin=0.00,
        ymax=3.5,
        grid=major,
        xlabel={Time steps},
        ylabel={Loss \(l(\theta)\)},
        xlabel style={at={(0.50,0.05)}},
        ylabel style={at={(0.06,0.50)}},
        xtick={0.00,20.00,...,100.00},
        ytick={0.00,0.5,...,3.5},
        log ticks with fixed point,
        legend pos=north east,
        legend cell align={left},
        legend style={fill opacity=0.6, draw opacity=1.0, text opacity=1.0, font=\small}
        ]

        \addplot[black, solid, thick, line width = 1pt, mark=triangle, mark size={3.0}, mark repeat=10, mark phase=10] 
            table [x=x_data, y=y_GD, col sep=comma]{./SigL1_d20k20.csv};
        
        \addplot[red, solid, thick, line width = 1pt, mark=square, mark size={3.0}, mark repeat=10, mark phase=10] 
            table [x=x_data, y=y_BinSearch, col sep=comma]{./SigL1_d20k20.csv};
        
        \addplot[cyan, solid, thick, line width = 1pt, mark=pentagon, mark size={3.0}, mark repeat=10, mark phase=10] 
            table [x=x_data, y=y_AdamDef, col sep=comma]{./SigL1_d20k20.csv};
        
        \addplot[blue, solid, thick, line width = 1pt, mark=o, mark size={3.0}, mark repeat=10, mark phase=10] 
            table [x=x_data, y=y_AdamNotDef, col sep=comma]{./SigL1_d20k20.csv};
        
        
        \addplot[yellow, solid, thick, line width = 1pt, mark=asterisk, mark size={3.0}, mark repeat=10, mark phase=10] 
            table [x=x_data, y=y_AdaDeltaNotDef, col sep=comma]{./SigL1_d20k20.csv};
        
        \addplot[orange, solid, thick, line width = 1pt, mark=diamond, mark size={3.0}, mark repeat=10, mark phase=10] 
            table [x=x_data, y=y_RMSprop, col sep=comma]{./SigL1_d20k20.csv};
        
       \addplot[brown, solid, thick, line width = 1pt, mark=+, mark size={3.0}, mark repeat=10, mark phase=10] 
            table [x=x_data, y=y_Adagrad, col sep=comma]{./SigL1_d20k20.csv};





    \end{axis}
\end{tikzpicture}   
  \caption{Sigmoid activation: d=20, k=20}
  \label{fig:sigd20k20}
\end{subfigure}
\caption{Single hidden layer network: Comparison with other optimization techniques}
\label{fig:L1optComp}
\end{figure*}
\revise{The proposed binary search algorithm outperforms all the other methods in all the cases as illustrated in Fig. \ref{fig:L1optComp}. (Kindly note that the legend provided in the first subplot holds for the all the figures and is not repeated for ease of viewing.) It should be noted that all the above methods (Adam, RMSProp, Adagrad, Adadelta and gradient descent with derived bound) only require a single evaluation of the optimization algorithm whereas the BinarySearch method is employed when multiple evaluations can be performed; for our experiments, we have considered 10 evaluations for the binary search method. However, we can see that the performance of the optimization algorithm Adam with its default learning rate is fairly poor as compared to the tuned version. Note that this tuning would also take up evaluations based on the search algorithm employed.\\
In the case of ReLU activation, we note that the derived bound itself outperforms all the other optimization methods. In case of the sigmoid activation function, the Adam optimizer with a learning rate of 0.01 exhibits faster convergence than the derived bound for \(d=10\) and \(k=10\); both Adam and RMSProp with 0.01 outperform the derived bound for \(d=20\) and \(k=20\). However, the choice of learning rate as 0.01 would require some tuning as the default rate is 0.001.}

\subsection{Two hidden layer networks}
\subsubsection{Comparison with HyperOpt}
For a two hidden layer network, we run GD for \(T=200\) epochs as it takes greater number of epochs to converge than the single hidden layer. Similar to the case of a single hidden layer, the fraction of experiments for which HyperOpt chooses divergent values for a network with ReLU activation is tabulated in Table \ref{tab:BinvsHyper2LReluDiv}.
\begin{table}[h]
    \centering
    \begin{tabular}{|c|c|c|c|c|c|}
        \hline
        {} & {} & {}& \multicolumn{3}{c|}{No. of evaluations}\\
        \hline
         \(d\) & \(k_1\) & \(k_2\) & 5 & 10 & 20 \\
         \hline
         5 & 3 & 2  & 0.07 & 0.02 & 0 \\
         10 & 5 & 3  &  0.29 & 0.23 & 0.05\\
         5 & 10 & 5  & 0.44 & 0.34 &  0.21\\
         10 & 10 & 10 & 0.77 & 0.5 & 0.44\\
         \hline
    \end{tabular}
    \caption{Fraction of times HyperOpt diverges for 2 hidden layer ReLU}
    \label{tab:BinvsHyper2LReluDiv}
\end{table}
We note that as the dimensions of the problem gets bigger, the number of experiments which return unusable (divergent) learning rates increases. For example, in the case of ReLU activation with \(k_1=10, k_2=10\), we obtain divergent learning rates for \(44\%\) of the experiments even after allowing 20 evaluations for HyperOpt. In case of sigmoid activation, it is again noted that there is no divergent behaviour. The fraction of the remaining experiments in which BinarySearch outperforms HyperOpt is tabulated in Table \ref{tab:BinvsHyper2L}. It is noticed that the proposed method overtakes the existing method at higher dimensions. Best-trace graphs for a two-hidden layer network resembles the graphs for a single hidden layer and are not produced due to lack of space.

\begin{table}[h]
    \centering
    \begin{tabular}{|c|c|c|c|c|c|c|c|c|}
        \hline
        {} & {} & {} & \multicolumn{3}{c|}{ReLU activation} & \multicolumn{3}{c|}{Sigmoid activation}\\
        \hline
        {} & {} & {}& \multicolumn{3}{c|}{No. of evaluations} & \multicolumn{3}{c|}{No. of evaluations}\\
        \hline
         \(d\) & \(k_1\) & \(k_2\) & 5 & 10 & 20 & 5 & 10 & 20\\
         \hline
         5 & 3 & 2 & 0.45 &  0.65 & 0.59 & 1 & 0.98 & 0.95 \\
         10 & 5 & 3  & 0.58 & 0.67 & 0.61& 1 & 0.99 & 0.96\\
         5 & 10 & 5 & 0.52 & 0.62 & 0.73 & 0.4 & 0.93 & 0.96\\
         10 & 10 & 10 & 0.56 & 0.66 & 0.86 &0 &1 & 1 \\
         \hline
    \end{tabular}
    \caption{Fraction of times the best value for BinarySearch outperforms HyperOpt for 2 hidden layer out of successful experiments}
    \label{tab:BinvsHyper2L}
\end{table}
\revise{For a two-layer network, we consider the architecture with \(d=5, k_1=3, k_2= 2\). This is chosen at random for study, and the constants over 100 experiments are tabulated in Table \ref{tab:comp3}; we notice similar trends for other architecture with different widths as well.}
\begin{table}[h]
    \centering
    \begin{tabular}{|c|c|c|c|c|c|c|}
    \hline
       Activation & Evals & Algorithm  &  Mean & Std. dev & Max & Min\\
    \hline     
     \multirow{4}{*}{ReLU} & \multirow{2}{*}{5} & BinarySearch & 0.431 & 0.435 & 3.287 & 0.102\\
     & & HyperOpt & 0.285 & 0.239 & 0.996 &0.015 \\ 
     \cline{2-7}
     & \multirow{2}{*}{20} & BinarySearch & 0.424 & 0.391 & 2.952 & 0.061\\
     & & HyperOpt & 0.308 & 0.218 & 0.915 & 0.001\\
    \hline
    \multirow{4}{*}{Sigmoid} & \multirow{2}{*}{5} & BinarySearch & 6.012 & 0.929 & 8.055 & 3.983\\ 
     & & HyperOpt & 0.677 &  0.222 & 0.999 & 0.122\\ 
     \cline{2-7}
     & \multirow{2}{*}{20} & BinarySearch & 18.656 & 6.662 & 32.364 & 5.849\\
     & & HyperOpt & 0.756 & 0.177 & 0.991 & 0.136 \\ 
    \hline
    \end{tabular}
   \caption{Variation of the chosen learning rates for two hidden layer network}
    \label{tab:comp3}
\end{table}

\revise{From Table \ref{tab:comp3}, we see that the average as well as the maximum learning rate chosen by BinarySearch is greater than HyperOpt. In our experiments, note that greater learning rate implies better convergence as we only consider non-divergent traces. Similar to the single hidden layer network, BinarySearch outperforms HyperOpt by a large margin in case of sigmoid activation, as it opts for learning rates greater than one.
}
\begin{figure*}[ht]
\begin{subfigure}{.5\textwidth}
  \centering
  \begin{tikzpicture}[thick]
    \begin{axis}[
        width=8cm,
        height=6cm,
        xmin=0,
        xmax=100,
        ymin=0.00,
        ymax=1.1,
        grid=major,
        xlabel={Time steps},
        ylabel={Loss \(l(\theta)\)},
        xlabel style={at={(0.50,0.05)}},
        ylabel style={at={(0.06,0.50)}},
        xtick={0.00,20.00,...,100.00},
        ytick={0.00,0.25,...,1.00},
        log ticks with fixed point,
        legend pos=north east,
        legend cell align={left},
        legend style={fill opacity=0.6, draw opacity=1.0, text opacity=1.0, font=\small}
        ]

        \addplot[black, solid, thick, line width = 1pt, mark=triangle, mark size={3.0}, mark repeat=10, mark phase=10] 
            table [x=x_data, y=y_GD, col sep=comma]{./ReluL2_5_3_2.csv};
        \addlegendentry{Derived bound};
        
        \addplot[red, solid, thick, line width = 1pt, mark=square, mark size={3.0}, mark repeat=10, mark phase=10] 
            table [x=x_data, y=y_BinSearch, col sep=comma]{./ReluL2_5_3_2.csv};
        \addlegendentry{Binary Search};
        
        \addplot[cyan, solid, thick, line width = 1pt, mark=pentagon, mark size={3.0}, mark repeat=10, mark phase=10] 
            table [x=x_data, y=y_AdamDef, col sep=comma]{./ReluL2_5_3_2.csv};
        \addlegendentry{Adam (default)};
        
        \addplot[blue, solid, thick, line width = 1pt, mark=o, mark size={3.0}, mark repeat=10, mark phase=10] 
            table [x=x_data, y=y_AdamNotDef, col sep=comma]{./ReluL2_5_3_2.csv};
        \addlegendentry{Adam (0.01)};
        
        
        \addplot[yellow, solid, thick, line width = 1pt, mark=asterisk, mark size={3.0}, mark repeat=10, mark phase=10] 
            table [x=x_data, y=y_AdaDeltaNotDef, col sep=comma]{./ReluL2_5_3_2.csv};
        \addlegendentry{Adadelta (0.01)};
        
        \addplot[orange, solid, thick, line width = 1pt, mark=diamond, mark size={3.0}, mark repeat=10, mark phase=10] 
            table [x=x_data, y=y_RMSprop, col sep=comma]{./ReluL2_5_3_2.csv};
        \addlegendentry{RMSprop (0.01)};
        
        \addplot[brown, solid, thick, line width = 1pt, mark=+, mark size={3.0}, mark repeat=10, mark phase=10] 
            table [x=x_data, y=y_Adagrad, col sep=comma]{./ReluL2_5_3_2.csv};
        \addlegendentry{Adagrad (0.01)};





    \end{axis}
\end{tikzpicture} 
  \caption{ReLU activation: d=5, \(k_1=3\), \(k_2=2\)}
  \label{fig:L2_relu_1}
\end{subfigure}
\begin{subfigure}{.5\textwidth}
  \centering
  \begin{tikzpicture}[thick]
    \begin{axis}[
        width=8cm,
        height=6cm,
        xmin=0,
        xmax=100,
        ymin=0.00,
        ymax=450,
        grid=major,
        xlabel={Time steps},
        ylabel={Loss \(l(\theta)\)},
        xlabel style={at={(0.50,0.05)}},
        ylabel style={at={(0.06,0.50)}},
        xtick={0.00,20.00,...,100.00},
        ytick={0.00,100.00,...,400.00},
        log ticks with fixed point,
        legend pos=north east,
        legend cell align={left},
        legend style={fill opacity=0.6, draw opacity=1.0, text opacity=1.0, font=\small}
        ]

        \addplot[black, solid, thick, line width = 1pt, mark=triangle, mark size={3.0}, mark repeat=10, mark phase=10] 
            table [x=x_data, y=y_GD, col sep=comma]{./ReluL2_10_10_10.csv};
        
        \addplot[red, solid, thick, line width = 1pt, mark=square, mark size={3.0}, mark repeat=10, mark phase=10] 
            table [x=x_data, y=y_BinSearch, col sep=comma]{./ReluL2_10_10_10.csv};
        
        \addplot[cyan, solid, thick, line width = 1pt, mark=pentagon, mark size={3.0}, mark repeat=10, mark phase=10] 
            table [x=x_data, y=y_AdamDef, col sep=comma]{./ReluL2_10_10_10.csv};
        
        \addplot[blue, solid, thick, line width = 1pt, mark=o, mark size={3.0}, mark repeat=10, mark phase=10] 
            table [x=x_data, y=y_AdamNotDef, col sep=comma]{./ReluL2_10_10_10.csv};
        
        
        \addplot[yellow, solid, thick, line width = 1pt, mark=asterisk, mark size={3.0}, mark repeat=10, mark phase=10] 
            table [x=x_data, y=y_AdaDeltaNotDef, col sep=comma]{./ReluL2_10_10_10.csv};
        
        \addplot[orange, solid, thick, line width = 1pt, mark=diamond, mark size={3.0}, mark repeat=10, mark phase=10] 
            table [x=x_data, y=y_RMSprop, col sep=comma]{./ReluL2_10_10_10.csv};
        
        \addplot[brown, solid, thick, line width = 1pt, mark=+, mark size={3.0}, mark repeat=10, mark phase=10] 
            table [x=x_data, y=y_Adagrad, col sep=comma]{./ReluL2_10_10_10.csv};





    \end{axis}
\end{tikzpicture}   
  \caption{ReLU activation: d=10, \(k_1=10\), \(k_2=10\)}
  \label{fig:L2_relu_2}
\end{subfigure}
\\
\begin{subfigure}{.5\textwidth}
  \centering
  \begin{tikzpicture}[thick]
    \begin{axis}[
        width=8cm,
        height=6cm,
        xmin=0,
        xmax=100,
        ymin=0.00,
        ymax=0.16,
        grid=major,
        xlabel={Time steps},
        ylabel={Loss \(l(\theta)\)},
        xlabel style={at={(0.50,0.05)}},
        ylabel style={at={(0.06,0.50)}},
        xtick={0.00,20.00,...,100.00},
        ytick={0.00,0.05,...,0.15},
        log ticks with fixed point,
        legend pos=north east,
        legend cell align={left},
        legend style={fill opacity=0.6, draw opacity=1.0, text opacity=1.0, font=\small}
        ]

        \addplot[black, solid, thick, line width = 1pt, mark=triangle, mark size={3.0}, mark repeat=10, mark phase=10] 
            table [x=x_data, y=y_GD, col sep=comma]{./SigL2_5_3_2.csv};
        
        \addplot[red, solid, thick, line width = 1pt, mark=square, mark size={3.0}, mark repeat=10, mark phase=10] 
            table [x=x_data, y=y_BinSearch, col sep=comma]{./SigL2_5_3_2.csv};
        
        \addplot[cyan, solid, thick, line width = 1pt, mark=pentagon, mark size={3.0}, mark repeat=10, mark phase=10] 
            table [x=x_data, y=y_AdamDef, col sep=comma]{./SigL2_5_3_2.csv};
        
        \addplot[blue, solid, thick, line width = 1pt, mark=o, mark size={3.0}, mark repeat=10, mark phase=10] 
            table [x=x_data, y=y_AdamNotDef, col sep=comma]{./SigL2_5_3_2.csv};
        
        
        \addplot[yellow, solid, thick, line width = 1pt, mark=asterisk, mark size={3.0}, mark repeat=10, mark phase=10] 
            table [x=x_data, y=y_AdaDeltaNotDef, col sep=comma]{./SigL2_5_3_2.csv};
        
        \addplot[orange, solid, thick, line width = 1pt, mark=diamond, mark size={3.0}, mark repeat=10, mark phase=10] 
            table [x=x_data, y=y_RMSprop, col sep=comma]{./SigL2_5_3_2.csv};
        
        \addplot[brown, solid, thick, line width = 1pt, mark=+, mark size={3.0}, mark repeat=10, mark phase=10] 
            table [x=x_data, y=y_Adagrad, col sep=comma]{./SigL2_5_3_2.csv};





    \end{axis}
\end{tikzpicture}  
  \caption{Sigmoid activation: d=5, \(k_1=3\), \(k_2=2\)}
  \label{fig:L2_sig_1}
\end{subfigure}
\begin{subfigure}{.5\textwidth}
  \centering
  \begin{tikzpicture}[thick]
    \begin{axis}[
        width=8cm,
        height=6cm,
        xmin=0,
        xmax=100,
        ymin=0.00,
        ymax=0.42,
        grid=major,
        xlabel={Time steps},
        ylabel={Loss \(l(\theta)\)},
        xlabel style={at={(0.50,0.05)}},
        ylabel style={at={(0.06,0.50)}},
        xtick={0.00,20.00,...,100.00},
        ytick={0.00,0.1,...,0.4},
        log ticks with fixed point,
        legend pos=north east,
        legend cell align={left},
        legend style={fill opacity=0.6, draw opacity=1.0, text opacity=1.0, font=\small}
        ]

        \addplot[black, solid, thick, line width = 1pt, mark=triangle, mark size={3.0}, mark repeat=10, mark phase=10] 
            table [x=x_data, y=y_GD, col sep=comma]{./SigL2_10_10_10.csv};
        
        \addplot[red, solid, thick, line width = 1pt, mark=square, mark size={3.0}, mark repeat=10, mark phase=10] 
            table [x=x_data, y=y_BinSearch, col sep=comma]{./SigL2_10_10_10.csv};
        
        \addplot[cyan, solid, thick, line width = 1pt, mark=pentagon, mark size={3.0}, mark repeat=10, mark phase=10] 
            table [x=x_data, y=y_AdamDef, col sep=comma]{./SigL2_10_10_10.csv};
        
        \addplot[blue, solid, thick, line width = 1pt, mark=o, mark size={3.0}, mark repeat=10, mark phase=10] 
            table [x=x_data, y=y_AdamNotDef, col sep=comma]{./SigL2_10_10_10.csv};
        
        
        \addplot[yellow, solid, thick, line width = 1pt, mark=asterisk, mark size={3.0}, mark repeat=10, mark phase=10] 
            table [x=x_data, y=y_AdaDeltaNotDef, col sep=comma]{./SigL2_10_10_10.csv};
        
        \addplot[orange, solid, thick, line width = 1pt, mark=diamond, mark size={3.0}, mark repeat=10, mark phase=10] 
            table [x=x_data, y=y_RMSprop, col sep=comma]{./SigL2_10_10_10.csv};
        
       \addplot[brown, solid, thick, line width = 1pt, mark=+, mark size={3.0}, mark repeat=10, mark phase=10] 
            table [x=x_data, y=y_Adagrad, col sep=comma]{./SigL2_10_10_10.csv};





    \end{axis}
\end{tikzpicture}   
  \caption{Sigmoid activation: d=10, \(k_1=10\), \(k_2=10\)}
  \label{fig:L2_sig_2}
\end{subfigure}
\caption{Two hidden layer network: Comparison with other optimization techniques}
\label{fig:L2optComp}
\end{figure*}
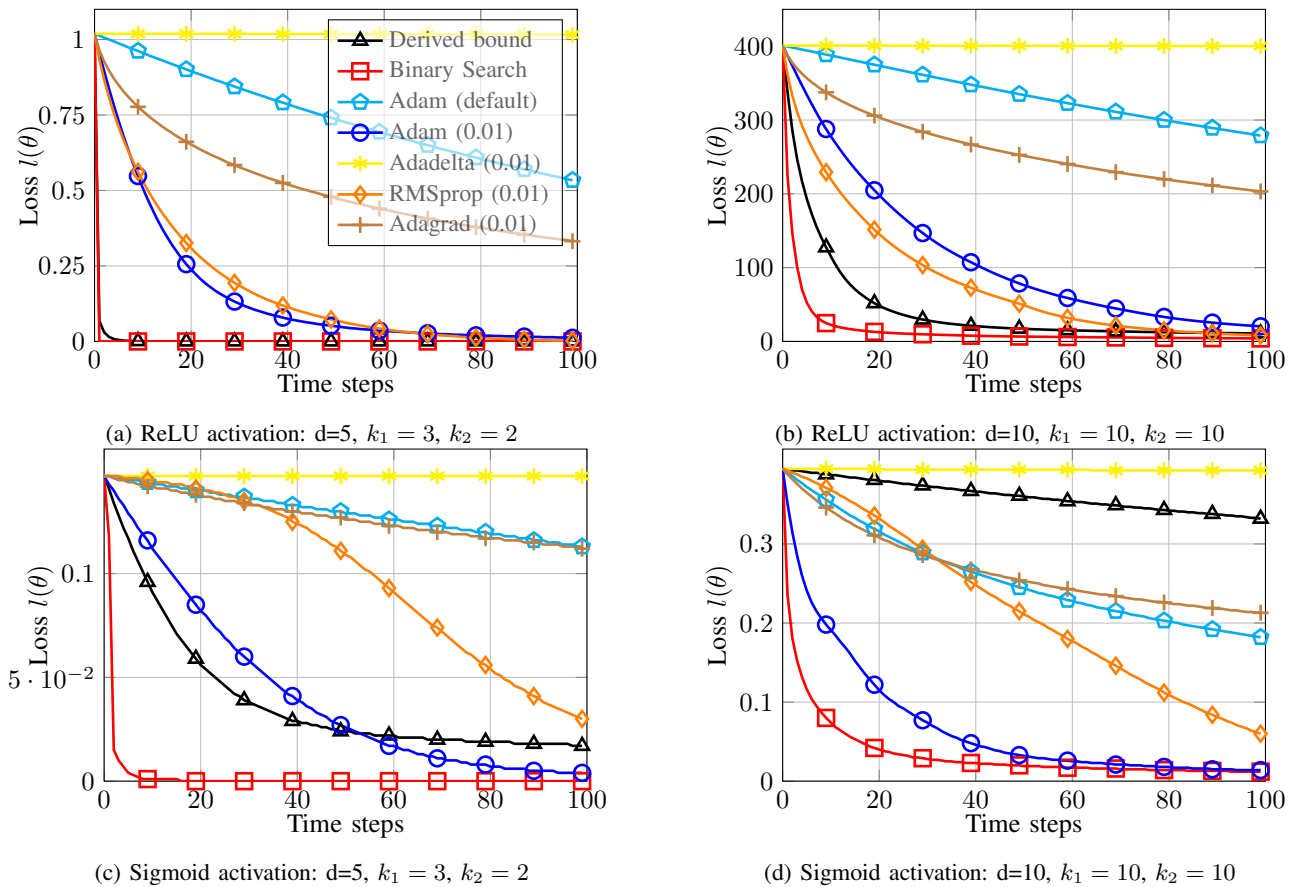

\subsubsection{Comparison with other optimization algorithms}
\revise{For the case of two layer networks, we see that the proposed binary search (using 10 evaluations) always results in faster convergence as compared to the other optimization algorithms. This is shown in Fig. \ref{fig:L2optComp}.
Similar to that of a single hidden layer network, for ReLU activation, we see that the derived bound outperforms the other optimization methods. However, for sigmoid activation, especially in higher network dimensions, the other algorithms perform better than the derived bound but worse than the proposed binary search. Although all the above experiments were performed with \(N=100\), the trend in performance does not change with change in \(N\) as both the loss function as well as the derived bound contain a factor of \(1/N\).} 
\vspace{-0.4cm}
\subsection{Remarks}
\subsubsection{Complexity of the algorithm}
\revise{The complexity of the proposed method as well as the comparative method, namely HyperOpt, is \(n \mathcal{C}(GD)\) where \(n\) is the number of evaluations and \(\mathcal{C}(GD)\) is the complexity of the plain vanilla gradient descent  algorithm for a fixed number of epochs. Note that both these tuning methods run the gradient descent algorithm during each evaluation for the same number of epochs. Therefore, both algorithms have the same complexity as long as they employ the same number of evaluations. Newer optimization methods such as Adam, Adagrad, RMSProp and Adadelta have greater algorithmic complexity than the traditional gradient descent algorithm as they involve more additive and multiplicative operations in order to maintain an adaptive step-size. Although the difference in computational complexities is not much among the adaptive algorithms, the trend in complexity is as follows: GD \(<\) Adagrad \(<\) RMSProp \(<\) Adam. \cite{wang2019optimization}.}
\subsubsection{Using the derived bound in a different search}
\added{One could ask if the derived bound can be used in HyperOpt or other existing popular hyper-parameter optimization algorithms itself. Though we can employ the derived bounds to restrict the search space of existing algorithms on one end of the interval, how to define the other end is still a question. For example, we note that for a neural network with sigmoid activation function, GD converges for learning rates greater than 1. Hence, the learning rate corresponding to the derived bound \((>1)\) may be set as the lower limit of the search interval; however, how to set the upper bound still remains a question. We believe that this is worth exploring in future work.}

	\section{Applications} \label{sec:appln}
Feedforward shallow networks are widely used in the context of resource allocation \cite{eisen2019learning}, wireless communication \cite{tacspinar2010back,huang2018deep}, financial predictions \cite{galeshchuk2016neural} and weather forecasting \cite{yadav2017wind}. In this section, we illustrate the utility of the proposed algorithm in \revise{three} specific applications.
\subsection{Channel Estimation in OFDM systems}
The use of neural network for channel estimation is advocated as traditional estimation methods such as Least Squares and MMSE suffer from lack of accuracy and high computational complexity respectively \cite{tacspinar2010back,jiang2019deep}. We now describe the architecture employed in \cite{tacspinar2010back}. A pilot-based channel estimation is considered. A single hidden layer with \(k\) neurons with sigmoid activation function is employed. The real and the imaginary parts of the received pilots are fed separately into the network and the corresponding channel impulses are estimated at the output. The output layer (with linear activation) has the same number of neurons as the input layer, say \(2M\) for estimating the channel response the real and imaginary parts of \(M\) sub-carriers. The component-wise sum of the squared difference between the estimated and actual channel response is the objective function to be minimized. The learning rate employed in the paper is 0.05 and is chosen through manual tuning, which usually involves searching through trial and error which is a laborious process. We now derive an upper bound on the gradient Lipschitz constant of the objective and apply Algorithm \ref{alg:BinSearch} to find the learning rate. \\
We follow the notation introduced in Section \ref{sec:2layer} where the weight matrix between the input and hidden layer is denoted by \(\bm{V}\) and the weight matrix between the hidden and output layer is denoted as \(\bm{W}\). Let the data points be denoted as \((\bm{x}(i), \bm{o}(i))\) for \(i=1,...N\). Each element of the output vector is denoted as \(o(i)_{l_2}\) where \(l_2 = 1,.. 2M\). The loss function is given by, 
\begin{equation}
    l(\bm{\theta}) = \dfrac{1}{2N} \sum_{i=1}^N \sum_{l_2 =1}^{2M} \left[\left(\sum_{l_1 = 1}^k \sigma(\bm{x}(i)\bm{^T V^{l_1}}) W_{l_1 l_2}  \right) - o(i)_{l_2} \right]^2.
\end{equation}
Note that, in this application, the architecture consists of multiple outputs nodes. Therefore, the result in Theorem \ref{theorem:optConvSig} cannot be used as it is. The bound on the gradient Lipschitz constant hence is derived for this specific case, and the bound is given by,
\begin{align}
    \alpha^* &\leq \dfrac{1}{N} \sum_{i=1}^N \bigg[\frac{k_1 k_2}{16} \beta^2 \norm{\bm{x}(i)}_{\infty} \norm{\bm{x}(i)}_{1} + \nonumber\\ 
    &\sum_{l_2 = 1}^{k_2} \left[(k_1 \beta - o_{l_2}) \frac{ \beta}{10} \norm{\bm{x}(i)}_{\infty} \norm{\bm{x}(i)}_{1} \right] +\nonumber\\
    &\frac{k_1 k_2}{4} \beta \norm{\bm{x}(i)}_{\infty} + \frac{k_1}{4} (k_1 \beta - \min_m o_{m}) \norm{\bm{x}(i)}_{\infty} \bigg].
\end{align}

\added{\textit{Sketch of Proof:} The elements of the Hessian matrix \(\nabla^2 l(\bm{\theta})\) are computed and the Gershgorin theorem (Theorem \ref{theorem:gershgorin}) is then applied to obtain the above result.}\\
Here, we consider a OFDM system with \(M=64\) sub-carriers where all the sub-carriers consist of the pilot symbol. The pilots are transmitted through the channel and received. All the simulations are performed in the frequency domain. It is assumed that the channel impulse responses are available for training. 
As done in \cite{tacspinar2010back}, the number of inputs and outputs to the neural network are \(2M\) and the number of neurons in the hidden layer are \(k=10\). For our simulations, we have considered a QPSK constellation and an SNR of 10dB. 

The learning rate chosen in the paper is a fixed learning rate 0.05. The loss corresponding to the fixed learning rate after \(T=100\) time steps is 0.068. The learning rate chosen by Algorithm \ref{alg:BinSearch} and the corresponding loss in tabulated in Table \ref{tab:app1}.
\begin{table}[h]
    \centering
    \begin{tabular}{|c|c|c|c|}
    \hline
         &  \multicolumn{3}{c|}{No. of evaluations}\\
    \hline     
     & 5& 10& 20 \\
     \hline
    Learning rate & 0.033 & 0.062 & 0.064\\
    \hline
    Loss&0.0004 & 3.53e-6 & 3.27 e-23\\
    \hline
    \end{tabular}
    \caption{Learning rates chosen and loss encountered for channel estimation by Algorithm \ref{alg:BinSearch}  }
    \label{tab:app1}
\end{table}
We can see that the proposed method finds a learning rate that is comparable to the one suggested by manual tuning with as low as 5 evaluations. We can also see that the loss that the algorithm converges to is lower than the loss arrived at by using 0.05 as the learning rate. 

\vspace{-0.4cm}
\subsection{Exchange rate prediction}
\begin{figure}[ht]
    \centering
    \resizebox{\linewidth}{!}{\begin{tikzpicture}[thick]
\begin{axis}[
        width=6cm,
        height=4.5cm,
        xmin=0,
        xmax=30,
        ymin=0.00,
        ymax=30,
        grid=major,
        xlabel={Time steps},
        ylabel={Loss \(l(\theta)\)},
        xlabel style={at={(0.50,0.05)}},
        ylabel style={at={(0.06,0.50)}},
        xtick={0.00,10.00,...,30.00},
        ytick={0.00,10.00,...,30.00},
        log ticks with fixed point,
        legend pos=north east,
        legend cell align={left},
        legend style={fill opacity=0.6, draw opacity=1.0, text opacity=1.0, font=\small}
        ]
        \pgfplotsset{every tick label/.append style={font=\small}}
        
        \addplot[red, solid, thick, line width = 1pt, mark=triangle, mark size={2.0}, mark repeat=5, mark phase=5] 
            table [x=x_data, y=y_BinarySearch, col sep=comma]{./app2.csv};
        \addlegendentry{BinarySearch};
        
        \addplot[blue, solid, thick, line width = 1pt, mark=square, mark size={2.0}, mark repeat=5, mark phase=5] 
            table [x=x_data, y=y_HyperOpt, col sep=comma]{./app2.csv};
        \addlegendentry{HyperOpt};

    \end{axis}
\end{tikzpicture}}
    \caption{Best trace comparison for exchange rate prediction network with 10 evaluations}
    \label{fig:app2}
\end{figure}
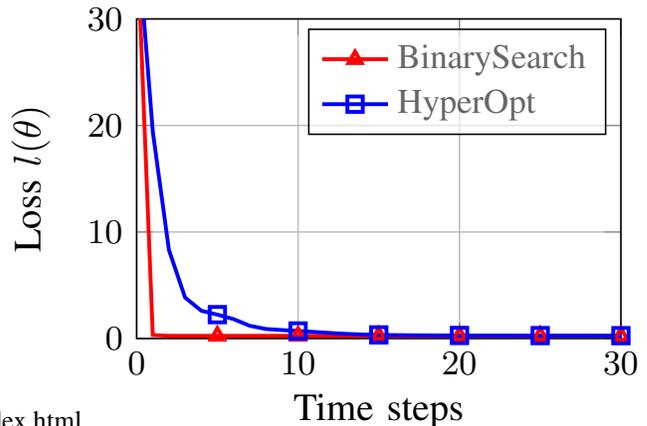

Neural networks are used in various aspects of finance such as debt risk assessment, currency prediction, business risk failure, etc. \cite{perez2017applicability}. Applications such as exchange rate prediction hold great importance in the economy. In \cite{galeshchuk2016neural}, a single hidden layer neural network is considered where the neurons employ the sigmoid activation function. In the mentioned work, prediction is done using daily, monthly or quarterly steps. For the sake of our demonstration, we consider the daily step prediction. The exchange rates for the previous \(d=5\) days are fed as the input to the network and the prediction for the next day is made. The architecture of the network is the same as the one demonstrated in Fig. \ref{fig:1Larch} with \(d=5\) input neurons, \(k=10\) neurons at the hidden layer and one output neuron. The data for the experiment is obtained from the website \url{http://www.global-view.com/forex-trading-tools/forex-history/index.html} as in \cite{galeshchuk2016neural}. \\
The data is organized as \((\bm{x}(i),y(i))\) for \(i=1,..N\) training samples; note that \(\bm{x}(i) \in \mathbb{R}^d\) represents the daily step (change in the exchange rate from the previous day) for the past five days and \(y(i)\) is the rate for the day (which is the quantity to be estimated). We implement \cite{galeshchuk2016neural} with a slight modification: the network proposed in the paper uses a threshold within every neuron which is also a parameter to be tuned; instead, in this implementation, we add a column of ones to the data to compensate for threshold. Hence, we have \(\bm{x}(i) \in \mathbb{R}^{d+1}\). We are justified in doing so as we would tune the weight vector corresponding to the \(d+1\)th input to the hidden layer instead of tuning the threshold. 
The loss function in \cite{galeshchuk2016neural} is given by, 
\begin{equation}
              l(\bm{w}) = \dfrac{1}{2N} \sum_{i = 1}^N \left(  \left(\sum_{j=1}^k \sigma(\bm{x}(i)^T \bm{w}^j) \right) - y(i) \right)^2,
\end{equation}
where \(\bm{w}\) denotes the weights of the network to be optimized and \(\sigma(.)\) denotes the sigmoid activation function. 
We note that the loss function is the similar to (\ref{eqn:loss1L}) and hence the bound derived in (\ref{eqn:1lsigfinal}) in Section \ref{sec:1layer} can be used. \\
The paper recommends GD as the optimization algorithm to be used; however, it does not recommend any tuning method for the learning rate for this application. We employ the BinarySearch method proposed in Algorithm \ref{alg:BinSearch} and tabulate the losses encountered after tuning the learning rate for \(T=500\) time steps in Table \ref{tab:app2}.
\begin{table}[h]
    \centering
    \begin{tabular}{|c|c|c|c|}
    \hline
         &  \multicolumn{3}{c|}{No. of evaluations}\\
    \hline     
     & 5& 10& 20 \\
    \hline
    BinarySearch & 0.254 & 0.2532 &0.253\\
    \hline
    HyperOpt &0.255 & 0.255 & 0.254 \\
    \hline
    \end{tabular}
    \caption{Loss encountered for exchange rate prediction by Algorithm \ref{alg:BinSearch} and HyperOpt }
    \label{tab:app2}
\end{table}
We note that the proposed method performs well as compared to HyperOpt using TPE and is able to achieve the optimal loss within a small number of iterations. \added{As it is noted that both the algorithms converge to similar losses, we wish to demonstrate the convergence graphs by plotting the best-trace graphs. From Fig. \ref{fig:app2}, we note that the proposed BinarySearch algorithm converges faster.}\\

\vspace{-0.4cm}
\subsection{Offset estimation in OFDM receivers}
\begin{figure}[ht]
  \centering
  \begin{tikzpicture}[thick]
    \begin{axis}[
        width=8cm,
        height=6cm,
        xmin=0,
        xmax=800,
        ymin=0.00,
        ymax=8,
        grid=major,
        xlabel={Time steps},
        ylabel={Loss \(l(\theta)\)},
        xlabel style={at={(0.50,0.05)}},
        ylabel style={at={(0.06,0.50)}},
        xtick={0.00,200.00,...,800.00},
        ytick={0.00,2.00,...,8.00},
        log ticks with fixed point,
        legend pos=north east,
        legend cell align={left},
        legend style={fill opacity=0.6, draw opacity=1.0, text opacity=1.0, font=\small}
        ]

        \addplot[black, solid, thick, line width = 1pt, mark=triangle, mark size={3.0}, mark repeat=100, mark phase=100] 
            table [x=x_data, y=y_GD, col sep=comma]{./CFO_est.csv};
        \addlegendentry{Derived bound};
        
        \addplot[red, solid, thick, line width = 1pt, mark=square, mark size={3.0}, mark repeat=100, mark phase=100] 
            table [x=x_data, y=y_BinSearch, col sep=comma]{./CFO_est.csv};
        \addlegendentry{Binary Search};
        
        \addplot[blue, solid, thick, line width = 1pt, mark=pentagon, mark size={3.0}, mark repeat=100, mark phase=100] 
            table [x=x_data, y=y_AdamDef, col sep=comma]{./CFO_est.csv};
        \addlegendentry{Adam - 0.001};
        
    \end{axis}
\end{tikzpicture}
  \caption{Learning curves for CFO estimation}
  \label{fig:cfo_est}
\end{figure}
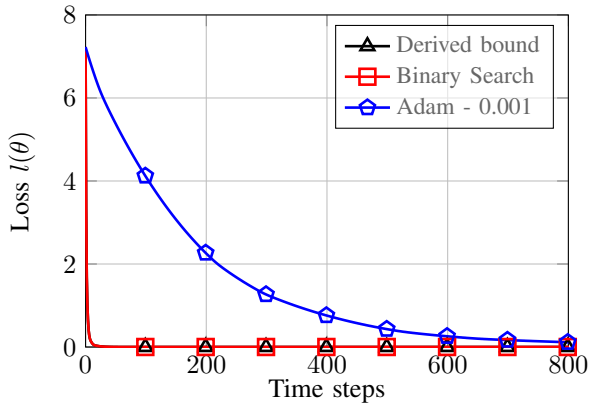

\revise{Recent work in \cite{ouameur2020model} uses neural network blocks for different purposes while designing an OFDM receiver such as synchronization. We focus on the estimation of the Carrier Frequency Offset(CFO). Shallow networks with restricted width are employed to reduce the computational complexity.\\
Simulation setup of \cite{ouameur2020model} is used and is briefly described below. The OFDM signal is generated as per IEEE 802.11 standard using 64 subcarriers and 4-equally spaced pilots where 16-QAM constellation is employed for modulation. The baseline model is derived using the estimate from the cyclic prefix (CP). A moving window of the CP estimates derived from \(N_{CFO}\) consecutive OFDM symbols serve as inputs to the shallow neural network. The estimate from the established preamble method developed by Moose \cite{moose1994technique} is used as the label for training.\\
The said shallow network is constructed with the same architecture in Fig.\ref{fig:1Larch} with ReLU activation; hence, the upper bound on the gradient Lipschitz constant is given in (\ref{eqn:1LReLUfinal}). It is established in \cite{ouameur2020model} that using a neural network to estimate the CFO as mentioned above results in better MSE than simply using the CP estimates or the preamble methods. We only verify if the proposed tuning method results in better learning curves than the optimization algorithm used in \cite{ouameur2020model}.\\
The authors employ the Adam optimization algorithm with the default learning rate of 0.001. We now compare this with the proposed method in Fig. \ref{fig:cfo_est}. It can be seen from the figure that using the derived bound as the learning rate of GD (as well as the binary search method) converges within a few initial epochs whereas Adam optimizer takes 800 epochs to converge to the same minimum. Note that although binary search requires more evaluations, GD with the derived learning rate requires just one iteration, like the Adam optimizer. This shows that our proposed method results in faster convergence.}

 \revise{In this section, we considered three popular applications in the communication and finance sector where shallow feedforward networks are used and demonstrated that the proposed method can be used effectively to tune the learning rate as compared to the state-of-the-art tuning algorithms.}
 
 \section{Concluding remarks}
\added{In this work, we proposed a theory-based approach for determining the learning rate for a shallow feedforward neural network. We derived the gradient Lipschitz constant for fixed architectures and developed a search algorithm that employs the derived bound to find a better learning rate while ensuring convergence. While the existing algorithms tune harder, i.e., employ higher number of evaluations in order to find a suitable learning rate, we can tune smarter by searching over an interval which is customized to the objective. When allowed the same number of evaluations, we demonstrated that the proposed method outperforms state-of-the-art methods such as HyperOpt in terms of convergence in both synthetic and real data.}

	 \appendices

  \section{Proof of Theorem \ref{theorem:optConv1LReLU}}
  \label{Appendix_A}
      As the function is doubly differentiable, the required constant is \(\alpha^* = \max_{\bm{w}} \lambda_{max}(\nabla^2 l(\bm{w}) )\).
     \begin{align}
         \nabla l(\bm{w}) &= \left( \sum_{j=1}^k s(\bm{x}^T \bm{w}^j) - y \right) 
         \begin{bmatrix} \mathbb{I}_{\{\bm{x}^T\bm{w}^1 \geq 0\}}  \bm{x}\\ \vdots \\ \mathbb{I}_{\{\bm{x}^T\bm{w}^k \geq 0\}}  \bm{x} \end{bmatrix} 
    \end{align}
    \begin{align}
         \nabla^2 l(\bm{w}) &=  \begin{bmatrix} \mathbb{I}_{\{\bm{x}^T\bm{w}^1 \geq 0\}} \bm{x} \\ \vdots \\ \mathbb{I}_{\{\bm{x}^T\bm{w}^k \geq 0\}}\bm{x} \end{bmatrix}  \begin{bmatrix} \mathbb{I}_{\{\bm{x}^T\bm{w}^1 \geq 0\}}  \bm{x}\\ \vdots \\ \mathbb{I}_{\{\bm{x}^T\bm{w}^k \geq 0\}} \bm{x} \end{bmatrix} ^T \nonumber \\
         &=  \bm{a(x,w}) \bm{a(x,w})^T
     \end{align}
     where 
        \begin{equation} \label{eqn:avec}
            \bm{a(x,w}) \triangleq \begin{bmatrix} \mathbb{I}_{\{\bm{x}^T\bm{w}^1 \geq 0\}}  \bm{x}& \hdots & \mathbb{I}_{\{\bm{x}^T\bm{w}^k \geq 0\}}  \bm{x} \end{bmatrix}^T.
        \end{equation}
       

    Although the ReLU function given by \(\max(0,x)\) is non-differentiable at \(x=0\), the work in \cite{li2017convergence} states that if the input is assumed to be from the Gaussian distribution, the loss function becomes smooth, and the gradient is well defined everywhere. The gradient is given by \(\mathbb{I}_{\{x \geq 0\}}\) where \(\mathbb{I}\) is the indicator function. By a similar argument, we consider the second derivative to be zero over the entire real line. Note that the Gaussian assumption is only to ensure that the derivative of the ReLU function is defined at \(x=0\) due to the smoothness for theoretical tractability. 
     The gradient Lipschitz constant is given by
     \begin{equation} \label{eqn:concOnePt}
         \alpha^* = \max_{\bm{w}} \lambda_{max}(\nabla^2 l(\bm{w}) )= \max_{\bm{w}} \lambda_{\max} ( \bm{a(x,w}) \bm{a(x,w})^T).
     \end{equation}
     
     We note that \(\bm{a(x,w}) \bm{a(x,w})^T\) is a rank-1 matrix and therefore, its only non-zero eigenvalue is given by \(\bm{a(x,w})^T \bm{a(x,w})= \norm{\bm{a(x,w})}^2\), which is also the maximum eigenvalue. Substituting in (\ref{eqn:concOnePt}),
     \begin{equation} \label{eqn:bound1}
         \alpha^* = \max_{\bm{w}} \norm{\bm{a(x,w})}^2.
     \end{equation}
     The norm is maximized when all the entries of the vector are non-zero, i.e., when all the indicators correspond to 1. Let us define 
      \begin{equation} \label{eqn:abar}
            \bm{\bar{a}(x)} \triangleq \begin{bmatrix} \bm{x}  & \hdots & \bm{x} \end{bmatrix}^T,
        \end{equation}
        which is a stack of the input vector repeated \(k\) times.
    Therefore, the required constant is given by 
     \begin{align}
         \alpha^* = \norm{\bm{\bar{a}(x)}}^2= k \norm{\bm{x}}^2.
     \end{align}
     In this case, we note that the derived constant for a single data point is not a bound, but the exact gradient Lipschitz constant and it is a function of the number of neurons, \(k\), and the norm of the input vector.

     \section{Proof of Lemma \ref{lemma:lem1}}
     \label{Appendix_B}
        
         The Rayleigh quotient of a Hermitian matrix \(\bm{A}\) and a non-zero vector \(\bm{g}\) is given by \(\frac{ \bm{g}^T \bm{A} \bm{g}}{ \bm{g}^T \bm{g}}\) and reaches the maximum eigenvalue when the vector \(\bm{g}\) is the eigen vector corresponding to the maximum eigenvalue \cite{van2007new}.
         \begin{equation}
             \lambda_{max}(\bm{A}) = \max_{\bm{g}: \norm{\bm{g}}=1} \bm{g}^T \bm{A} \bm{g} ,
         \end{equation}
         Also, observe that for any other vector of unit norm \(\bm{h} \neq \bm{g}\),
         \begin{equation} \label{eqn:mat}
             \bm{g}^T \bm{A} \bm{g} > \bm{h}^T \bm{A} \bm{h}.
         \end{equation}
         In the following proof, denoting \(\bm{x}(i)\) as \(\bm{x_i}\) and the principal eigen vectors of \(\left( \sum_{i=1}^N  \bm{\bar{a}(x_i)} \bm{\bar{a}(x_i)} ^T \right)\), \( \left(\bm{a(x_i,w}) \bm{a(x_i,w})^T  \right)\) and \(\left( \sum_{i=1}^N \bm{a(x_i,w}) \bm{a(x_i,w})^T  \right) \) as \(\bm{\bar{g}}, \bm{g_i}\) and \(\bm{\hat{g}}\) respectively, 
         
         \begin{align*}
             \lambda_{max}\left( \sum_{i=1}^N  \bm{\bar{a}(x_i)} \bm{\bar{a}(x_i)} ^T \right) &= \bm{\bar{g}}^T \left( \sum_{i=1}^N  \bm{\bar{a}(x_i)} \bm{\bar{a}(x_i)} ^T \right) \bm{\bar{g}} \\
             &= \sum_{i=1}^N \bm{\bar{g}}^T \left( \bm{\bar{a}(x_i)} \bm{\bar{a}(x_i)} ^T \right) \bm{\bar{g}} \\
             & \geq \sum_{i=1}^N \bm{g_i}^T \left( \bm{\bar{a}(x_i)} \bm{\bar{a}(x_i)} ^T \right) \bm{g_i}^T .
        \end{align*}
        Note that as \(\left(\bm{a(x_i,w}) \bm{a(x_i,w})^T  \right)\) is a rank-1 matrix, the principal eigen vector is given by \( \bm{g_i} = \bm{a(x_i,w})\). Hence, 
        \begin{align} \label{eqn:step2}
            &\sum_{i=1}^N \bm{g_i}^T \left( \bm{\bar{a}(x_i)} \bm{\bar{a}(x_i)} ^T \right) \bm{g_i}^T \nonumber\\
             &=  \sum_{i=1}^N \bm{a(x_i,w})^T  \left( \bm{\bar{a}(x_i)} \bm{\bar{a}(x_i)} \right) \bm{a(x_i,w}).
        \end{align}
        Considering each term in the summation,
        \begin{align*}
            &\bm{a(x_i,w})^T  \left( \bm{\bar{a}(x_i)} \bm{\bar{a}(x_i)} ^T \right) \bm{a(x_i,w}) \\
            &\quad \quad=  \bigg(\bm{a(x_i,w})^T \bm{\bar{a}(x_i)} \bigg) \bigg( \bm{\bar{a}(x_i})^T \bm{a(x_i,w}) \bigg)\\
            &\quad \quad= \bigg(\sum_{j=1}^k \mathbb{I}_{\{\bm{x_i^T w^j \geq0\}}} \bm{x_i^T x_i} \bigg)\bigg( \sum_{j=1}^k \mathbb{I}_{\{\bm{x_i^T w^j \geq0\}}} \bm{x_i^T x_i} \bigg)\\
            &\quad \quad= \bigg(\sum_{j=1}^k \mathbb{I}^2_{\{\bm{x_i^T w^j \geq0\}}} \bm{x_i^T x_i} \bigg)\bigg( \sum_{j=1}^k \mathbb{I}^2_{\{\bm{x_i^T w^j \geq0\}}} \bm{x_i^T x_i} \bigg)\\
            &\quad \quad= \bigg(\sum_{j=1}^k \mathbb{I}_{\{\bm{x_i^T w^j \geq0\}}} \bm{x_i^T} \mathbb{I}_{\{\bm{x_i^T w^j \geq0\}}} \bm{ x_i} \bigg)\\
            &\quad \quad \quad \quad \bigg(\sum_{j=1}^k \mathbb{I}_{\{\bm{x_i^T w^j \geq0\}}} \bm{x_i^T} \mathbb{I}_{\{\bm{x_i^T w^j \geq0\}}} \bm{ x_i}  \bigg)\\
            &\quad \quad =  \bm{a(x_i,w})^T  \left( \bm{a(x_i,w}) \bm{a(x_i,w}) ^T \right) \bm{a(x_i,w}).
        \end{align*}
        Using this result in (\ref{eqn:step2}),
        \begin{align*}
            &\sum_{i=1}^N \bm{a(x_i,w})^T  \left( \bm{\bar{a}(x_i)} \bm{\bar{a}(x_i)} ^T \right) \bm{a(x_i,w}) \\
             & \quad \quad= \sum_{i=1}^N \bm{a(x_i,w})^T  \left( \bm{a(x_i,w}) \bm{a(x_i,w}) ^T \right) \bm{a(x_i,w}) \\
             &\quad \quad \geq \sum_{i=1}^N \bm{\hat{g}}^T  \left( \bm{a(x_i,w}) \bm{a(x_i,w}) ^T \right)  \bm{\hat{g}}\\
             &\quad \quad =  \bm{\hat{g}}^T  \left(\sum_{i=1}^N \bm{a(x_i,w}) \bm{a(x_i,w}) ^T \right)  \bm{\hat{g}} \\
             &\quad \quad = \lambda_{max}\left( \sum_{i=1}^N \bm{a(x_i,w}) \bm{a(x_i,w})^T  \right).
         \end{align*}
     Hence proved.
     
     \vspace{-0.4cm}
     \section{Proof of Theorem \ref{theorem:2layerSigConv}}
     \label{Appendix_C}
      
     The loss function is doubly differentiable, and hence, 
     \begin{equation}
         \alpha^* = \max_{\bm{\theta}} \lambda_{max}(\nabla^2 l(\bm{\theta}) ).
     \end{equation}
     The first-order partial derivatives are computed as follows,
         \begin{align}
         \dfrac{\partial l(\bm{\theta})}{\partial \bm{\theta}} &= A \dfrac{\partial A}{\partial \bm{\theta}}\\
         \dfrac{\partial A}{\partial \bm{\theta}} &= \begin{bmatrix}
            \left(\sum_{l_2 = 1}^{k_2} \left[ q_{l_2} \sigma(\bm{x}^T \bm{V}^{1})(1-\sigma(\bm{x}^T \bm{V}^{1})) W_{1 l_2} \bm{x} \right] \right)\\
            \vdots \\
            \left(\sum_{l_2 = 1}^{k_2} \left[ q_{l_2} \sigma(\bm{x}^T \bm{V}^{k_1})(1-\sigma(\bm{x}^T \bm{V}^{k_1})) W_{k_1 l_2} \bm{x} \right] \right)\\
            q_1 \sigma(\bm{x}^T \bm{V}^{1}) \\
            \vdots \\
            q_1 \sigma(\bm{x}^T \bm{V}^{k_1}) \\
             q_2 \sigma(\bm{x}^T \bm{V}^{1}) \\
            \vdots \\
            q_2 \sigma(\bm{x}^T \bm{V}^{k_1})\\
            \vdots\\
            q_{k_2} \sigma(\bm{x}^T \bm{V}^{k_1}) 
         \end{bmatrix} \label{eqn:da_sigm}
     \end{align} 
    We define the following terms, where \(q_a\) is the first derivative and \(q_a'\) is the second derivative of \(\sigma \left(\sum_{l_1=1}^{k_1} ( \sigma(\bm{x}^T \bm{V} ^{l_1}) W_{l_1 a}) \right)\) and then compute the elements of the Hessian matrix.
     \begin{align}
         A &\triangleq \left( \sum_{l_2 = 1}^{k_2} \sigma\left(\sum_{l_1 = 1}^{k_1} \sigma(\bm{x}^T \bm{V}^{l_1}) W_{l_1 l_2} \right) - y \right), \\
         q_a &\triangleq \sigma \left(\sum_{l_1=1}^{k_1} ( \sigma(\bm{x}^T \bm{V} ^{l_1}) W_{l_1 a}) \right) \nonumber \\
         & \qquad \quad \left( 1 - \sigma \left(\sum_{l_1=1}^{k_1} ( \sigma(\bm{x}^T \bm{V} ^{l_1}) W_{l_1 a}) \right) \right),
    \end{align}
     \begin{align}
         q'_a &\triangleq \sigma \left(\sum_{l_1=1}^{k_1} ( \sigma(\bm{x}^T \bm{V} ^{l_1}) W_{l_1 a}) \right) \nonumber \\
         & \qquad \quad \left( 1 - \sigma \left(\sum_{l_1=1}^{k_1} ( \sigma(\bm{x}^T \bm{V} ^{l_1}) W_{l_1 a}) \right) \right)\nonumber \\
         & \qquad \quad \left( 1 - 2\sigma \left(\sum_{l_1=1}^{k_1} ( \sigma(\bm{x}^T \bm{V} ^{l_1}) W_{l_1 a}) \right) \right).\\
         \dfrac{\partial^2 l(\bm{\theta})}{\partial W_{ij} \partial W_{i'j'}} &= \left( q_j \sigma(\bm{x}^T \bm{V}^{i}) \right) \left( q_{j'} \sigma(\bm{x}^T \bm{V}^{i'}) \right) \nonumber \\
         & \quad \quad +A \sigma(\bm{x}^T \bm{V}^{i}) \sigma(\bm{x}^T \bm{V}^{i'}) q'_j \mathbb{I}_{\{ j=j'\}} \label{eqn:pd1}
    \end{align}
     \begin{align}     
         \dfrac{\partial^2 l(\bm{\theta})}{\partial \bm{V}^{i} \partial \bm{V}^{i'}} &= \left(\sum_{l_2 = 1}^{k_2} \left[ q_{l_2} \sigma(\bm{x}^T \bm{V}^{i})(1-\sigma(\bm{x}^T \bm{V}^{i})) W_{i l_2} \bm{x} \right] \right) \nonumber\\
         & \left(\sum_{l_2 = 1}^{k_2} \left[ q_{l_2} \sigma(\bm{x}^T \bm{V}^{i'})(1-\sigma(\bm{x}^T \bm{V}^{i'})) W_{i' l_2} \bm{x} \right] \right)^T \nonumber \\
         &+ A \sigma(\bm{x}^T \bm{V}^{i})(1-\sigma(\bm{x}^T \bm{V}^{i})) 
         \sigma(\bm{x}^T \bm{V}^{i'})\nonumber\\
         &(1-\sigma(\bm{x}^T \bm{V}^{i'})) 
         \left[ \sum_{l_2 = 1}^{k_2} q'_{l_2} W_{i l_2}W_{i' l_2} \right] \bm{xx^T}\nonumber\\
         &+ A \left[ \sum_{l_2 = 1}^{k_2} q_{l_2} W_{i l_2} \right] \sigma(\bm{x}^T \bm{V}^{i})(1-\sigma(\bm{x}^T \bm{V}^{i})) \nonumber\\
         &(1-2\sigma(\bm{x}^T \bm{V}^{i})) \mathbb{I}_{\{ i=i'\}} \bm{xx^T} \label{eqn:pd2}\\ 
        \dfrac{\partial^2 l(\bm{\theta})}{\partial \bm{V}^{i} \partial W_{i'j'}} &=  \sum_{l_2 = 1}^{k_2} \left[q_{l_2} \sigma(\bm{x}^T \bm{V}^{i})(1-\sigma(\bm{x}^T \bm{V}^{i})) W_{i l_2} \right]\nonumber\\
        &q'_j \sigma(\bm{x}^T \bm{V}^{i'}) \bm{x}^T + A \sigma(\bm{x}^T \bm{V}^{i})(1-\sigma(\bm{x}^T \bm{V}^{i}))\nonumber \\
        &\bigg[\sum_{l_2 = 1}^{k_2} ( q_{l_2} \mathbb{I}_{\{i=i',l_2=j'\}} + W_{i l_2} q'_{l_2} \sigma(\bm{x}^T \bm{V}^{i'}) \bigg] \bm{x}^T
        \label{eqn:pd3}
    \end{align}
    \begin{align}  
        \dfrac{\partial^2 l(\bm{\theta})}{\partial W_{ij}\partial \bm{V}^{i'} } &= q_j \sigma(\bm{x}^T \bm{V}^{i})\nonumber\\
        &\sum_{l_2 = 1}^{k_2} \left[q_{l_2} \sigma(\bm{x}^T \bm{V}^{i'})(1-\sigma(\bm{x}^T \bm{V}^{i'})) W_{i' l_2} \bm{x} \right]\nonumber\\
        &+A q_j \sigma(\bm{x}^T \bm{V}^{i})(1-\sigma(\bm{x}^T \bm{V}^{i}))  \mathbb{I}_{i=i'} \bm{x}\nonumber\\
        &+ A (\sigma(\bm{x}^T \bm{V}^{i'}))^2 (1-\sigma(\bm{x}^T \bm{V}^{i'})) q'_jW_{i'j} \bm{x}\label{eqn:pd4}
    \end{align}
    It is observed that the elements of the Hessian matrix depends on the values of the parameters in \(\bm{\theta}\) (through \(A\)) unlike the case with a single hidden layer in which the parameters only appeared as indicators. As the maximization is over \(\bm{\theta}\), the elements of the matrix \(\bm{V}\) and \(\bm{W}\) can be scaled up arbitrarily and the obtained upper bound will be infinity, which is a trivial upper bound. To avoid this, it is assumed that the magnitude of the weights are restricted; i.e., \(|\theta_i|< \beta \quad \forall i\). The Hessian matrix can be written in the following form:
    \begin{align}
        \nabla^2 l(\bm{\theta}) &= \left(\dfrac{d A}{d \bm{\theta}} \right) \left(\dfrac{d A}{d \bm{\theta}} \right)^T + \bm{M},
    \end{align}
    where the first terms in all the second order partial derivative elements (given in (\ref{eqn:pd1}) - (\ref{eqn:pd4})) are accounted for in \(\left(\frac{d A}{d \bm{\theta}} \right) \left(\frac{d A}{d \bm{\theta}} \right)^T\). The rest of the additive terms are represented by the matrix \(\bm{M}\).
     Applying Weyl's inequality (i.e., (\ref{eqn:weyl})), 
    \begin{equation} \label{eqn:2lsiglambdamx}
        \lambda_{max}(\nabla^2 l(\bm{\theta})) \leq  \lambda_{max}\left( \left(\dfrac{d A}{d \bm{\theta}} \right) \left(\dfrac{d A}{d \bm{\theta}} \right)^T \right)+  \lambda_{max}(\bm{M}).
    \end{equation}
    We note that the first term in the above equation is a rank one matrix and has a maximum eigenvalue of \(\norm{\frac{d A}{d \bm{\theta}}}^2\). The required gradient Lipschitz constant is obtained by maximizing (\ref{eqn:2lsiglambdamx}) over all values of \(\bm{\theta}\) and is given by 
    \begin{equation} \label{eqn:maximizingSig}
        \max_{\bm{\theta}} \lambda_{max}(\nabla^2 l(\bm{\theta})) \leq  \max_{\bm{\theta}} \norm{\dfrac{d A}{d \bm{\theta}}}^2 + \max_{\bm{\theta}}  \lambda_{max}(\bm{M}) 
    \end{equation}
     Focusing on the first term in (\ref{eqn:maximizingSig}), the vector \(\frac{d A}{d \bm{\theta}}\) consists of \(k_1 k_2\) terms of the form \(q_{(.)} \sigma(.)\) and \(k_1\) terms of the form \(\sum_{l_2 = 1}^{k_2} q_{l_2} \nabla \sigma(\bm{x}^T\bm{V^{a}}) W_{a l_2} \bm{x}\) where \(a=1,...,k_1\). Recall from (\ref{eqn:func}) that \(\sigma(.) \leq 1\) and from (\ref{eqn:func1stder}) that \(q_{(.)} \leq \frac{1}{4}\). Therefore,
     \begin{equation} \label{eqn:maxfirstterm}
         \max_{\bm{\theta}} \norm{\dfrac{d A}{d \bm{\theta}}}^2 = k_1\left(\frac{k_2 \beta \norm{\bm{x}}}{16}\right)^2 + \frac{k_1 k_2}{16}
     \end{equation}
    where \(\beta = \max_i \theta_i\).
     We now focus on the second additive term in (\ref{eqn:maximizingSig}). To bound the maximum eigenvalue of \(\bm{M}\), the Gershgorin's theorem (stated in Theorem \ref{theorem:gershgorin}) is employed. Considering the terms in (\ref{eqn:pd1}) - (\ref{eqn:pd4}) that are not included in \(\left(\frac{d A}{d \bm{\theta}} \right) \left(\frac{d A}{d \bm{\theta}} \right)^T\), we bound the maximum row sum over all possible values of \(\bm{\theta}\).
     The row sum can be computed in one of two possible ways considering elements from (a) \(\dfrac{\partial^2 l(\bm{\theta})}{\partial \bm{V}^{i} \partial \bm{V}^{i'}}\) and \(\dfrac{\partial^2 l(\bm{\theta})}{\partial \bm{V}^{i} \partial W_{i'j'}}\) 
         or (b) \(\dfrac{\partial^2 l(\bm{\theta})}{\partial W_{ij}\partial \bm{V}^{i'} }\) and \( \dfrac{\partial^2 l(\bm{\theta})}{\partial W_{ij} \partial W_{i'j'}}\).\\
     The maximum value taken by \(A\) is \( |k_2 -y|\) as the sigmoid function has a maximum value of one. Recall that \(q_a\) is a first derivative and \(q_a'\) is a second derivative of the sigmoid function. Using the bounds on derivatives stated in (\ref{eqn:func}) - (\ref{eqn:func2ndder}), 

     \begin{align}
        & \max_{\bm{\theta}}  \lambda_{max}(\bm{M}) \leq |k_2 -y|  \max\bigg( \frac{1}{10} +  \left[\frac{1}{4} + \frac{\beta}{10}\right] \frac{ k_1 \norm{\bm{x}}_1}{4}, \nonumber \\
         & \left[\frac{1}{4} + \frac{\beta}{10}\right] \frac{k_2 \norm{\bm{x}}_{\infty}}{4} +\left[\frac{\beta}{1000}  + \frac{1}{4}\right] k_1 k_2 \beta \norm{\bm{x}}_1 \norm{\bm{x}}_{\infty} \bigg) \label{eqn:M}
     \end{align}
     where the first argument in the maximization corresponds to case (a) and the second argument corresponds to case (b) of computing the row sum.
     Combining (\ref{eqn:maxfirstterm}) and (\ref{eqn:M}), 
     \begin{align}
          \alpha^* &\leq \max\Bigg( \frac{1}{10} +  \left[\frac{1}{4} + \frac{\beta}{10}\right] \frac{ k_1 \norm{\bm{x}}_1}{4},  \left[\frac{1}{4} + \frac{\beta}{10}\right] \frac{k_2 \norm{\bm{x}}_{\infty}}{4}\nonumber \\
         & +\left[\frac{\beta}{1000}  + \frac{1}{4}\right] k_1 k_2 \beta \norm{\bm{x}}_1 \norm{\bm{x}}_{\infty} \Bigg)  |k_2 -y| \nonumber \\
         &
         + k_1\left(\frac{k_2 \beta \norm{\bm{x}}}{16} \right)^2 +  \frac{k_1 k_2}{16} 
     \end{align}
      
\section{Proof of Theorem 6}
     \label{Appendix_D}

     The aim is to find the gradient Lipschitz constant of \(l(\bm{\theta})\). For a doubly differentiable function, the required constant is given by 
     \begin{equation}
         \alpha^* = \max_{\bm{\theta}} \lambda_{max}(\nabla^2 l(\bm{\theta}) ).
     \end{equation}
     In order to find the Hessian, we initially find the first-order partial derivatives:
        \begin{align}
         \dfrac{\partial l(\bm{\theta})}{\partial \bm{\theta}} &= A \dfrac{\partial A}{\partial \bm{\theta}}
         \end{align}
         
         \begin{align}
         \dfrac{\partial A}{\partial \bm{\theta}} &= \begin{bmatrix}
            \left(\sum_{l_2 = 1}^{k_2} \left[ q_{l_2} \mathbb{I}_{\{\bm{x}^T \bm{V}^{1} \geq 0\}} W_{1 l_2} \bm{x} \right] \right)\\
            \vdots \\
            \left(\sum_{l_2 = 1}^{k_2} \left[ q_{l_2} \mathbb{I}_{\{\bm{x}^T \bm{V}^{k_1} \geq 0\}} W_{k_1 l_2} \bm{x} \right] \right)\\
            q_1 s(\bm{x}^T \bm{V}^{1}) \\
            q_1 s(\bm{x}^T \bm{V}^{2})  \\
            \vdots \\
            q_1 s(\bm{x}^T \bm{V}^{k_1}) \\
            \vdots \\
            q_{k_2} s(\bm{x}^T \bm{V}^{k_1}) \label{eqn:da}
         \end{bmatrix}
     \end{align} 
     where
     \begin{align}
         A &\triangleq \left( \sum_{l_2 = 1}^{k_2} s\left(\sum_{l_1 = 1}^{k_1} s(\bm{x}^T \bm{V}^{l_1}) W_{l_1 l_2} \right) - y \right) \\
         q_a &\triangleq \mathbb{I}_{\{\sum_{l_1=1}^{k_1} ( s(\bm{x}^T \bm{V}^{l_1}) W_{l_1 a}) \geq 0 \}}.
    \end{align}
 
     \imp{Similar to one-hidden layer ReLU case, we assume that the gradients of \(q_a\) with respect to \(W_{ij}\) and \(\bm{V}^{i}\) are \(0\) and \(\bm{0}\) respectively.}
     Now, the second-order partial derivatives are derived. 
     \begin{align}
         \dfrac{\partial^2 l(\bm{\theta})}{\partial W_{ij} \partial W_{i'j'}} &= \left( q_j s(\bm{x}^T \bm{V}^{i}) \right) \left( q_{j'} s(\bm{x}^T \bm{V}^{i'}) \right)\\
         \dfrac{\partial^2 l(\bm{\theta})}{\partial \bm{V}^{i} \partial \bm{V}^{i'}} &= \left(\sum_{l_2 = 1}^{k_2} \left[ q_{l_2} \mathbb{I}_{\{\bm{x}^T \bm{V}^{i} \geq 0\}} W_{i l_2} \bm{x} \right] \right) \nonumber\\
         & \qquad \left( \sum_{l_2 = 1}^{k_2} \left[ q_{l_2} \mathbb{I}_{\{\bm{x}^T \bm{V}^{i'} \geq 0\}} W_{i' l_2} \bm{x} \right] \right)^T 
    \end{align}
    \begin{align}
        \dfrac{\partial^2 l(\bm{\theta})}{\partial \bm{V}^{i} \partial W_{i'j'}} &= A q_{j'} \mathbb{I}_{\{\bm{x}^T \bm{V}^{i'} \geq 0\}} \mathbb{I}_{\{i = i'\}}\bm{x}^T + \nonumber \\
        & \left(\sum_{l_2 = 1}^{k_2} \left[ q_{l_2} \mathbb{I}_{\{\bm{x}^T \bm{V}^{i} \geq 0\}} W_{i l_2} \bm{x}^T \right] \right) \left(q_{j'} s(\bm{x}^T \bm{V}^{i'}) \right) \label{eqn:dvdw}\\
        \dfrac{\partial^2 l(\bm{\theta})}{\partial W_{ij}\partial \bm{V}^{i'} } &= A q_{j} \mathbb{I}_{\{\bm{x}^T \bm{V}^{i} \geq 0\}} \mathbb{I}_{\{i = i'\}}\bm{x} + \nonumber \\
        & \left(\sum_{l_2 = 1}^{k_2} \left[ q_{l_2} \mathbb{I}_{\{\bm{x}^T \bm{V}^{i'} \geq 0\}} W_{i' l_2} \bm{x}\right] \right) \left(q_{j} s(\bm{x}^T \bm{V}^{i}) \right). \label{eqn:dwdv}
    \end{align}
     Note that the Hessian is a square matrix of dimension \(k_1(d+k_2) \times k_1(d+k_2)\). On putting the Hessian matrix together, it is observed that the Hessian can be written as a sum of two matrices as given below
    \begin{align}
        \nabla^2 l(\bm{\theta}) &= \left(\dfrac{d A}{d \bm{\theta}} \right) \left(\dfrac{d A}{d \bm{\theta}} \right)^T + \bm{M},
    \end{align}
    where \(\bm{M}\) is a matrix with all the elements as zero except for the additional elements corresponding to \(\frac{\partial^2 l(\bm{\theta})}{\partial W_{ij}\partial \bm{V}^{i'} }\) and \(\frac{\partial^2 l(\bm{\theta})}{\partial \bm{V}^{i'} \partial W_{ij}}\) where \(i=i'\). The main diagonal elements of the matrix are always zero and it is also symmetric; there are \(2d k_1 k_2\) non-zero elements in the matrix. 
    
    Using Weyl's inequality stated in (\ref{eqn:weyl}),
    \begin{align}
        \lambda_{max}(\nabla^2 l(\bm{\theta})) &\leq  \lambda_{max}\left( \left(\dfrac{d A}{d \bm{\theta}} \right) \left(\dfrac{d A}{d \bm{\theta}} \right)^T \right)+  \lambda_{max}(\bm{M}) \\
        &= \norm{\dfrac{d A}{d \bm{\theta}}}^2 +  \lambda_{max}(\bm{M}). \label{eqn:some}
    \end{align}
    
    The maximum eigenvalue of the matrix \(\bm{M}\) can be bounded using the Brauer’s Ovals of Cassini bound (stated in Theorem \ref{theorem:Brauer}). 
     \begin{align} \label{eqn:alpha4}
         \lambda_{max}(\bm{M}) &\leq \max_{i \neq j} \bigg(\dfrac{m_{ii} + m_{jj}}{2} +  \nonumber \\
         & \hspace{1cm} \sqrt{(m_{ii} - m_{jj})^2 + R_i(\bm{M}) R_j(\bm{M})}  \bigg) 
     \end{align}
     where \(R_i(\bm{M}) = \sum_{i \neq j} |m_{ij}|\). 
     It is noted that all diagonal elements are always zero and multiple rows have similar row sums. Therefore, the bound reduces to 
     \begin{equation} \label{eqn:alpha4_2}
         \lambda_{max}(\bm{M}) \leq \max_{i }  R_i(\bm{M})
     \end{equation}
    This is the same as the Gershgorin's bound obtained for the matrix \(\bm{M}\). Note that the elements of the matrix \(\bm{M}\) are the first terms in (\ref{eqn:dvdw}) and (\ref{eqn:dwdv}) corresponding to the case when \(i=i'\). The structure of the matrix \(\bm{M}\) is such that the maximum row sum can be computed in one of two ways: \( A k_2\) times the maximum element of vector \(\bm{x}\), or \(A\) times the sum of elements of \(\bm{x}\). Therefore, while maximizing over \(\theta\), the maximum row sum of \(\bm{M}\) is given by 
     \begin{equation}
         \max_{i}  R_i(\bm{M}) = \max( A k_2 |\bm{x}|_{\infty}, A |\bm{x}|_1),
     \end{equation}
     where \(|\bm{x}|_{\infty}= \max_i \bm{x}_i\) and \(|\bm{x}|_{1}= \sum_i \bm{x}_i\).
     
    We can write (\ref{eqn:some}) as 
    \begin{equation} 
        \lambda_{max}(\nabla^2 l(\bm{\theta})) \leq \norm{\dfrac{d A}{d \bm{\theta}}}^2 +  \max( A k_2 |\bm{x}|_{\infty}, A |\bm{x}|_1).
    \end{equation}
    
    To obtain the desired bound on the gradient Lipschitz constant, we maximize over all possible values of \(\bm{\theta}\) to obtain,
    \begin{equation} \label{eqn:maximizing}
        \max_{\bm{\theta}} \lambda_{max}(\nabla^2 l(\bm{\theta})) \leq  \max_{\bm{\theta}} \norm{\dfrac{d A}{d \bm{\theta}}}^2 + \max_{\bm{\theta}}  \lambda_{max}(\bm{M}).
    \end{equation}
    
    The first term is an outer product of vectors (matrix of rank 1) and hence, the eigenvalue is given by their inner product. The vector \(\frac{d A}{d \bm{\theta}}\) consists of \(k_1(d+k_2)\) terms each with an indicator, an element from \(\theta\) and the input vector. Recall that to avoid arbitrary scaling of the derived bound, we impose the following restriction that \(|\theta_i| \leq \beta \qquad \forall i\). Therefore, 
    \begin{equation} \label{eqn:firstterm}
        \max_{\bm{\theta}} \norm{\dfrac{d A}{d \bm{\theta}}}^2 = k_1(d+k_2) \beta^2 \norm{\bm{x}}^2.
    \end{equation}

      To maximize the second term in (\ref{eqn:maximizing}), we note that the scalar term \(A\) is a sum of \(k_1k_2\) combinations of product of two weight parameters with the data vector \(\bm{x}\). The maximum value that the scalar \(A\) can take is denoted by \(A_{max} = k_1 k_2 \beta^2 \norm{\bm{x}} -y\). Therefore, the second term is maximized as 
      \begin{equation} \label{eqn:secondterm}
          \max_{\bm{\theta}}  \lambda_{max}(\bm{M}) = \max( (A_{max} k_2 |\bm{x}|_{\infty}, A_{max} |\bm{x}|_1),
      \end{equation}
      where \(A_{max} = k_1 k_2 \beta^2 \norm{\bm{x}} -y\). Combining (\ref{eqn:maximizing}), (\ref{eqn:firstterm}) and (\ref{eqn:secondterm}), we obtain
     \begin{equation} \label{eqn:final2l1ip}
         \alpha^* \leq k_1(d+k_2) \beta^2 \norm{\bm{x}}^2 + \max( (A_{max} k_2 |\bm{x}|_{\infty}, A_{max} |\bm{x}|_1).
     \end{equation}
     An upper bound on the gradient Lipschitz constant for a two hidden layer ReLU network is derived. 
    \cut{We note that the derived upper bound in (\ref{eqn:final2l1ip}) is a function of the network parameters \(k_1\) and \(k_2\) and the input parameters \(d\) and \(\bm{x}\).}
    
	\bibliographystyle{IEEEtran}
	\bibliography{IEEEabrv,arxiv_rev.bib}
	
\end{document}